%% file: paper.tex
\documentclass[Afour,sageh,times]{sagej}

\input{packages}
\setcitestyle{authoryear,open={(},close={)}}
\begin{document}
\input{title}

\graphicspath{ {./images/} }
\input{intro}
\input{prelim}
\input{approach}
\input{pivot-for-inference}
\input{results}
\input{conclusion}

\section{Declaration of Conflicting Interest}
The authors declare that there is no conflict of interest.

\section{Funding}
This research was partially supported by the Israel Science Foundation (ISF), by the Israel Ministry of Science and Technology (MOST), and by a donation from the Zuckerman Fund to the Technion Center for Machine Learning and Intelligent Systems (MLIS).

\bibliography{../../../References/refs}

\pagebreak
\appendix
\graphicspath{ {./images/} }
\input{appendix-obs}
\input{appendix-models}

\end{document}

%% file: packages.tex
\newcommand{\changed}[1]{{\color{black} #1}}
\newcommand{\changedRemoved}[1]{{\color{black} #1}}

\usepackage{amsmath}
\usepackage{centernot} 
\usepackage{amsfonts}
\usepackage{color}
\usepackage[dvipsnames]{xcolor}
\usepackage[font=small,labelfont=bf]{caption} 
\usepackage{multirow,multicol,tabularx}
\usepackage{url}
\usepackage{verbatim} 
\usepackage[T1]{fontenc} 

\usepackage{mathtools}
\DeclarePairedDelimiter\abs{\lvert}{\rvert}%
\DeclarePairedDelimiter\norm{\lVert}{\rVert}%
\makeatletter
\let\oldabs\abs
\def\abs{\@ifstar{\oldabs}{\oldabs*}}
\let\oldnorm\norm
\def\norm{\@ifstar{\oldnorm}{\oldnorm*}}
\makeatother

\usepackage{bm} 
\newcommand{\prob}{\mathbb{P}}
\DeclareMathOperator*{\argmax}{argmax}

\DeclareMathOperator*{\expect}{\mathbb{E}}
\newcommand{\inv}{\mathcal{I}nv}
\newcommand{\ninv}{\neg\mathcal{I}nv}
\newcommand{\entropy}{\mathrm H}

\newcommand{\pivot}{\mathtt{PIVOT}}

\usepackage [english]{babel}
\usepackage [autostyle, english = american]{csquotes}
\MakeOuterQuote{"}

\usepackage{graphicx}
\usepackage{epstopdf}

\usepackage[linesnumbered,ruled,vlined]{algorithm2e}
\makeatletter
\newcommand{\nosemic}{\renewcommand{\@endalgocfline}{\relax}}
\newcommand{\dosemic}{\renewcommand{\@endalgocfline}{\algocf@endline}}
\let\oldnl\nl
\newcommand{\nonl}{\renewcommand{\nl}{\let\nl\oldnl}}
\makeatother

\usepackage{setspace}

\usepackage{tikz}
\usetikzlibrary{shapes,arrows}
\usetikzlibrary{chains}

\usepackage{amsthm}

\theoremstyle{definition}

\theoremstyle{remark}

\usepackage{floatrow}
\usepackage{wrapfig}

\usepackage{subcaption}
\usepackage{empheq}

\usepackage[export]{adjustbox}

\usepackage{dblfloatfix}

\usepackage{authblk}

\usepackage{array}
\newcolumntype{?}{!{\vrule width 1.5pt}}
\usepackage{makecell}

%% file: title.tex
\title{Efficient Belief Space Planning in High-Dimensional State Spaces using PIVOT: Predictive Incremental Variable Ordering Tactic}

\author{Khen Elimelech\affilnum{1} and Vadim Indelman\affilnum{2}}
\runninghead{Elimelech and Indelman}
\affiliation{\affilnum{1}Robotics and Autonomous Systems Program, Technion –- Israel Institute of Technology.\\
             \affilnum{2}Department of Aerospace Engineering, Technion –- Israel Institute of Technology}
\corrauth{Khen Elimelech,\\ Technion, Haifa 3200003, Israel.}
\email{\texttt{khen@technion.ac.il}}

\begin{abstract}
In this work, we examine the problem of online decision making under uncertainty, which we formulate as planning in the belief space. Maintaining beliefs (i.e., distributions) over high-dimensional states (e.g., entire trajectories) was not only shown to significantly improve accuracy, but also allows planning with information-theoretic objectives, as required for the tasks of active SLAM and information gathering. Nonetheless, planning under this "smoothing" paradigm holds a high computational complexity, which makes it challenging for online solution. Thus, we suggest the following idea: before planning, perform a standalone state variable reordering procedure on the initial belief, and "push forwards" all the predicted loop closing variables. Since the initial variable order determines which subset of them would be affected by incoming updates, such reordering allows us to minimize the total number of affected variables, and reduce the computational complexity of candidate evaluation during planning. We call this approach PIVOT: Predictive Incremental Variable Ordering Tactic. Applying this tactic can also improve the state inference efficiency; if we maintain the PIVOT order after the planning session, then we should similarly reduce the cost of loop closures, when they actually occur. To demonstrate its effectiveness, we applied PIVOT in a realistic active SLAM simulation, where we managed to significantly reduce the computation time of both the planning and inference sessions. The approach is applicable to general distributions, and induces no loss in accuracy.
\end{abstract}

\keywords{Decision making under uncertainty, belief space planning, sparse systems, factor graphs, active SLAM.}

\maketitle

%% file: intro.tex
\section{Introduction}
\subsection{\changedRemoved{Background}}
Autonomous agents and robots deployed in the real world must be able to handle uncertainty, in order to achieve reliable and robust performance. For example, when solving tasks such as localization, mapping, or manipulation, mobile robots should often estimate their own state in the world, while accounting for their inherently noisy actuation, and only relying on noisy sensing or partial observability. They should also be able to propagate this estimate efficiently and sequentially, to allow long term autonomous operation. An important instance of this problem is Simultaneous Localization And Mapping (SLAM) \citep{Cadena16tro}, in which we wish to infer an agent's trajectory, while mapping its surroundings, \emph{in real time}. Such problems are especially challenging, as the estimated state is high-dimensional. 

Further, to allow real autonomous operation, once a robust state estimate is achieved, the agent should address the task of planning under uncertainty. This fundamental problem has long been a major interest of the robotics and Artificial Intelligence (AI) research communities. The goal in this task is, given the uncertainty-aware state estimate, to "actively" determine the \emph{future} course of action for the agent, in order to achieve a certain objective. To make a reliable decision, we should predict the development of uncertainty into the future, considering the different candidate actions or policies, and their possible outcomes. \linebreak Thus, while state inference is challenging for real-time solution, planning is exponentially more demanding.

The focus of this work is to allow computationally efficient planning under uncertainty, with a practical focus on "active SLAM"; yet, before formulating this problem and its solutions, it is important to first introduce the state inference problem, on which it relies.

\subsection{\changedRemoved{Sequential Probabilistic State Inference}}
In this problem, given a set of probabilistic constraints or noisy measurements, we would like to find the best, i.e., the maximum-a-posteriori (MAP), state estimate. To be able to express our confidence in the current estimate, and to allow fusion of information arriving from different sources, we typically maintain a \emph{belief} --  a posterior distribution over the state, given the available information. By utilizing Bayes'~rule, we are able to update ("propagate") the belief through time-steps, at the arrival of new information, in what is known as "Bayesian Inference". Broadly, we can divide the belief-based inference methods into two categories: filtering, in which we maintain the belief over the present agent pose (and possibly other variables of interest); and smoothing, in which we maintain a belief over a high-dimensional state, containing also past poses.

Generally, smoothing approaches enjoy a better accuracy and robustness, as maintaining the belief over the entire trajectory allows us to: (1) perform "loop closures", and update the estimate of past poses, when re-observing scenes; (2) perform re-linearization of past constraints, given the updated estimate, and, by such, further improve accuracy; (3)~understand topological properties of the constraint graph. For these reasons, this estimation paradigm is an appropriate choice for solving (e.g.) the SLAM problem. The application of this smoothing paradigm to SLAM was established by \cite{Dellaert06ijrr}, as "Smoothing and Mapping (SAM)". There, the authors suggested to represent the belief using a Probabilistic Graphical Model (PGM), the factor graph, which describes the topology of constraints (factors) \citep{Koller09book,Dellaert17foundations}.\linebreak
If the constraint models are linearized, and their noise is Gaussian, it was shown that the state inference can be performed using the upper triangular "square root matrix"~$\bm R$ of the system's information, given via the QR or Cholesky factorizations \citep{Golub96}.

In sequential problems, as time progresses, new constraints, and possibly new variables, are added to the system; then, the belief and its square root matrix shall be updated. However, propagation of beliefs, has, at worst, cubical complexity in terms of the state size, and it thus suffers from the "curse of dimensionality" \citep{Bellman57book}. State-of-the-art SAM approaches, such as iSAM2 \citep{Kaess12ijrr}, and SLAM++ \citep{Ila17ijrr}, partially alleviate this concern, by recognizing that the belief's square root matrix can be incrementally updated, starting from the first variable to receive a new constraint, according to the variable order.

\subsection{Belief Space Planning}
Following the previous discussion, we can now confidently examine the problem of online planning and decision making under uncertainty \citep{Kochenderfer15book}, which is the actual interest of this work. When solving this problem, we wish to select (online) an action sequence or a policy, out of a set of possible candidates, which we predict to maximize some expected objective. This problem is often modeled as (a variation of) the Markov Decision Process (MDP), or Partially Observable MDP (POMDP) \citep{Kaelbling98ai,Thrun05book}. While it might be theoretically possible to approximate the expected objectives by sampling and propagating \emph{state} particles along each predicted trajectory \citep{Thrun00nips,Stachniss05rss}, achieving a sufficient representation of a belief over the high-dimensional state would require an outstanding amount of such particles, making this approach unfeasible. We hence focus here only on parametric belief propagation, as previously discussed. Also, examining the belief space, instead of the state space, as considered in standard PO/MDPs, allows us to examine belief-based policies, and belief-based information-theoretic reward functions. This paradigm is thus useful for solving tasks such as information gathering, and active SLAM \citep{Stachniss04iros,Kim14ijrr}, where we wish to reduce the uncertainty over the entire maintained trajectory and/or feature map.

Thus, to evaluate the candidates, and select the optimal one, we shall predictively propagate the belief (usually) multiple time-steps into the future, according to (possibly) different hypotheses for each candidate; in other words, we shall solve multiple "parallel" state inference sessions, while predicting the future belief development, as if we were to follow each candidate. Essentially, this process means performing a forward search in the belief space, and we thus refer to this problem as Belief Space Planning (BSP) \citep{Bonet00aips}. Unfortunately, BSP is computationally challenging for online solution, and reducing its cost is of great importance. 

\subsection{\changed{Contribution}}
The previous discussion leads us to the goal of this work -- reducing the computational complexity of this planning problem, by optimizing the process of belief update required for candidate evaluation. As explained, during sequential inference, the order of variables in the belief determines which variables will be affected by an upcoming update. Potentially, we wish the amount of these affected variables to be minimal; yet, when facing such an update, it is not possible to manipulate the belief or the order of variables to make variables "unaffected", as doing so would inherently make the manipulated variables affected. 

However, during planning, we face a unique situation, in which the same initial belief faces \emph{multiple} different updates "in parallel", representing predicted hypotheses for different candidates actions. In this case, we suggest a simple yet effective idea: before performing these updates, perform a precursory standalone variable reordering procedure, and "push forwards" all loop closing variables, in order to minimize the total number of affected variables by these updates. We call this approach PIVOT: Predictive Incremental Variable Ordering Tactic. This approach is inherently complementary to various planning methods, as we only optimize the belief \emph{representation}. This also means that the approach does not comprise the accuracy of the belief nor the quality of solution.

Also, while PIVOT is intended to benefit the efficiency of planning, the tactic often benefits also the efficiency of state inference. If the optimized order is maintained \emph{after} planning, then the predicted loop closing variables would remain "forward" in the variable order. Thus, if our prediction coincides with the ground truth during execution, we should similarly reduce the cost of loop closures during inference, when they actually occurs. Further, this tactic is highly relevant when considering sequential re-planning, as the PIVOT order can be incrementally updated before each planning session. 

PIVOT was first introduced in our previous publication \citep{Elimelech19isrr}. In this article, we vastly extend and improve our analysis, by presenting new scalable ("multi-class") and fill-aware variants. We also show that it is applicable for more general scenarios, including non-Gaussian and non-linear systems; planning with multiple hypotheses per candidate (i.e., without relying on "maximum likelihood observations"); and general objective functions.

\subsection{\changed{Article Structure}}
The article is structured as follows:
we begin with an extensive overview and formulation of the state inference and belief space planning problems, in the context of active SLAM (Section~\ref{sec:prelim}). Afterwards, we present the novel concept of PIVOT and its algorithmic variations (Section~\ref{sec:pivot}). Although PIVOT aims to contribute primarily to planning, we follow by explaining its evident effects on state inference and re-planning (Section~\ref{sec:influence-on-inf}). We then present an experimental demonstration of the approach in a realistic active SLAM simulation (Section~\ref{sec:results}), where we managed to substantially improve the solution efficiency, with no sacrifice in accuracy. We finish with a discussion on the experimental results (Section~\ref{sec:discussion}), and some concluding remarks (Section~\ref{sec:conclusion}).

To keep the explanation clear, in our approach description, we assume that each candidate policy is matched with a single predicted hypothesis during planning; however, this is not an inherent assumption of PIVOT. An extended analysis describing the more general case is brought in Appendix~\ref{sec:appendix-general-obs}. Further, for conciseness, the ideas in the article are presented in the practical context of linear(ized) and Gaussian systems, in which the inference can be described through manipulation of matrices; however, the ideas presented in this article are applicable to general distributions, where inference can be described through manipulation of probabilistic graphical models (PGMs). In Appendix~\ref{sec:appendix-general-models} we extend the discussion, and explain how and why PIVOT is still applicable in such scenarios.

\subsection{\changedRemoved{Comparison to Additional Related Work}}
Most works which consider planning in the belief space, typically focus on methods for sampling the space, in order to alleviate the search for the optimal candidate \citep{Prentice09ijrr,Patil14wafr,AghaMohammadi18tro}; such methods only rely on belief filtering (i.e., considering a belief over a single pose), and not on the more accurate and general belief smoothing, as we do. This is not surprising, since, as explained, the computational cost of this problem can be exceptionally high (due to the "curse of dimensionality"); this provides further motivation to our approach, which tries to tackle this exact issue.

In fact, our previous work \citep{Elimelech18arxiv} presented another attempt in achieving this goal, through belief sparsification, rather than belief reordering. Since such sparsification intrinsically requires a variable reordering step, the two methods are essentially related. Still, they are logically different as sparsification conveys \emph{approximation}, while reordering conveys a change of \emph{representation}. Further, both methods are "scalable", though in different regards: sparsification -- by selecting more variable for spasification; and PIVOT -- by dividing the variables into more classes. Thus, although the most basic form of each approach ("action consistent sparsification" and $\pivot_1$) can have a similar effect on the updates, each approach enjoys it own unique strengths. For example, we can sparsify also "involved" (as to be explained) variables, and, by such, sacrifice accuracy for additional performance; on the other hand, with PIVOT, we do not have to discard the modified belief, and, by such, improve the inference process, and efficiently update the order during re-planning. Also, with PIVOT, unlike sparsification, we are not limited to information theoretic objectives.

The work by \cite{Kopitkov17ijrr,Kopitkov19ijrr} also focused on reducing the computational complexity of planning with high-dimensional beliefs. It showed that when relying on the information gain as an information-theoretic objective for planning, we do not need to explicitly propagate the belief, in order to derive the objective values; instead, we can utilize the matrix determinant lemma, to efficiently update the determinant of the initial belief's covariance matrix. Of course, our method does not limit the objective function used in planning, nor the type of belief distribution (which was, in their case, strictly Gaussian).

Similarly to us, another related work \citep{Farhi17icra} also considers a system-wide view, which jointly examines both the (belief space) planning and the inference processes. There, the authors suggested to re-use calculation from planning, during the following state inference sessions, when applying the selected action. We note that while our approach is meant to mainly benefit the planning (and as a "by-product" also the state inference), the intention of that approach is to solely benefit the inference process (as a "by-product" of the planning). Further, while both approaches lead to a reduction in the computational effort required in inference time, their approach requires maintenance of an external cache; in contrast, our method reaches this goal implicitly, with simply modifying the belief representation, and with no explicit intervention in the inference process.

One work, by \cite{Chaves16iros}, similarly examined variable ordering in the context of planning, to allow reuse of calculations when examining similar candidate actions or between similar planning sessions. This scheme is conceptually different from the one presented here, as that reordering occurs during the evaluation of the candidates (instead of it being a precursory step), and also requires external Bayes tree caching. Also, while their results are indeed impressive, that work is significantly more restricted than the one we propose here: they consider a problem-specific scenario (active SLAM), while we do not; they assume a certain belief structure ("pose graph"), a certain belief distribution (Gaussian), and certain topological properties of the candidate paths, while we make no assumptions of this nature; they also do not account for the possibility of multiple predicted hypotheses per candidate, while we do; and, finally, they only consider planning with predefined control sequences, while we also allow incrementally-evaluated policies. Besides this attempt, we are not aware of other variable ordering approaches designed specifically for planning, as we presented here.

Nevertheless, we recognize that variable ordering tactics (or heuristics) are used extensively in the solution of the state inference problem. Due to the equivalence between graphs and matrices, this concern lies in the intersection of sparse matrix algebra and graph theory, as recognized in SAM and our formulation \citep{Kepner11book}. As explained, to solve the inference problem, we need to find the factorized representation of the belief, given either as the "square root matrix" $\bm R$ (derived from the QR of Cholseky factorizations), or, more generally, an "eliminated" Bayesian network. The amount of non-zero entries in $\bm R$, which is known as its "fill-in", reflects the computational cost of this factorization, and relies heavily on the order of state variables. Thus, to reduce this cost, we often rely on fill-reducing variable reordering. Although finding the optimal fill-reducing order is NP-complete \citep{Yannakakis81}, various heuristics -- variations of the minimum degree algorithm, such as COLAMD \citep{Davis04toms}, or graph partitioning techniques, such as nested dissection \citep{Gilbert86nm} -- provide empirically good results. Problem-specific heuristics, e.g., for SLAM \citep{Agarwal12iros,Krauthausen06rss}, which exploit the known state topology, are also applicable.

We recall that in sequential estimation, at each time-step, we shall appropriately update the belief factorization, in order to refine our estimate. To efficiently keep up with such updates, the aforementioned incremental inference techniques \citep{Kaess12ijrr,Ila17ijrr} apply such fill-reducing orders incrementally, only on the subset of affected variables, during the partial re-factorization.  Thus, in sequential estimation, variable reordering is usually not applied globally (considering all the variables), and not as a standalone procedure. It should be mentioned that in an earlier version of the incremental smoothing and mapping algorithm (iSAM, by  \cite{Kaess08tro}), application of a fill-reducing order was indeed suggested in periodic standalone steps; yet, this approach was later neglected, as it was not cost-efficient, in comparison to the incremental (though possibly sub-optimal) application of such orders during updates. Nonetheless, we should clarify that although the suggested PIVOT order is fill-aware, it is not intended to be fill-reducing, but only to benefit the planning process. Also, unlike such ordering tactics, which arrange the variables according to the current belief topology, we uniquely suggest to order the variables based on its predicted future development.

%% file: prelim.tex
\section{\changedRemoved{Preliminaries and Problem Definition} \label{sec:prelim}}
\subsection{\changedRemoved{Probabilistic State Inference \label{sec:prelim-infer}}}
Let us consider a sequential stochastic optimization process. In such a process, we wish to maintain an estimate of a state vector of random variables, given a stream of probabilistic constraints over these variables.

For example, in the context of mobile robots, we may consider a Simultaneous Localization and Mapping (SLAM) process, which we describe as follows: at time-step $k$, an agent at pose $x_{k-1}$ transitions to pose $x_k$, using a control $u_k$, and then collects an observation $z_k$. The agent's state $\mathcal{X}_k \doteq \{x_0,\dots,x_k\} \cup \mathcal{L}_k$ is comprised of its entire trajectory up to that point, alongside (optionally) a collection of additional external variables $\mathcal{L}_k$, such as the positions of observed landmarks. We assume that the pose transition and the observations comply to the following models, respectively:
\begin{eqnarray}
\label{eq:model-trans}
x_k &=& g_k(x_{k-1}, u_k) + w_{k},\\
\label{eq:model-obs}
z_{k} &=& h_k(\mathcal{X}^o_k) + v_{k},
\end{eqnarray}
where $\mathcal{X}^o_k$ represents the subset of the state variables which are observed ("involved in the observation"); $g_k$ and $h_k$ are deterministic functions; and $w_k$ and $v_k$ are independent random variables, representing stochastic noise. We may also assume access to prior probabilistic knowledge on the initial~pose, given as the constraint
\begin{eqnarray}
\label{eq:model-prior}
x_0 &=& \overline{x_0} + \bm v_{0},
\end{eqnarray} 
where $\overline{x_0}$ is an initial estimate, and the random variable $v_0$ represents stochastic noise.

At each time-step, we wish to derive the posterior distribution over the state, given the controls and observations taken until that time, known as the agent's \emph{belief}:
\begin{equation}
b_k(\mathcal{X}_k) \doteq \prob(\mathcal{X}_k \mid \mathcal{U}_k, \mathcal{Z}_k),
\end{equation}
where $\mathcal{U}_k \doteq \left\{u_1,\dots,u_k\right\}$ and $\mathcal{Z}_k \doteq \left\{z_1,\dots,z_k\right\}$.
Inferring the belief over the state, or over a subset of its variables, is known as "probabilistic state inference". Maintaining the belief over the agent's entire trajectory is known as "smoothing", in comparison to "filtering", where we maintain the belief only over the most recent pose. Most often, from this belief, we are interested in finding the maximum-a-posteriori (MAP) state estimate \mbox{$ \mathcal{X}^*_k \doteq \argmax_{\mathcal{X}_k} b_k(\mathcal{X}_k)$} ("MAP inference").

To do so, we begin by applying Bayes' rule and the Markov assumption on the belief, which allows us to conveniently factorize it into the following product:
\begin{equation}
\label{eq:belief-factor-product}
b_k(\mathcal{X}_k) \propto f^\text{prior}_{\overline{x_0}}(x_0) \cdot \prod_{i=1}^k f^\text{motion}_{u_i}(x_{i-1},x_{i}) \prod_{i=1}^k f^\text{obs}_{z_i}(\mathcal{X}^o_k),
\end{equation}
where
\begin{eqnarray}
f^\text{motion}_{u_k}(x_{k-1},x_{k}) &\doteq & \prob(x_{k+1} \mid x_k, u_{k+1}),\\
f^\text{obs}_{z_k}(\mathcal{X}^o_k) &\doteq & \prob(z_k \mid \mathcal{X}^o_k), \\
f^\text{prior}_{\overline{x_0}}(x_0) &\doteq & \prob(x_0).
\end{eqnarray}
Each of these \emph{factors} represents a probabilistic constraint on the state variables, matching the aforementioned prior, transition, and observation models.

\begin{figure}[t]
\includegraphics[width=0.75\columnwidth]{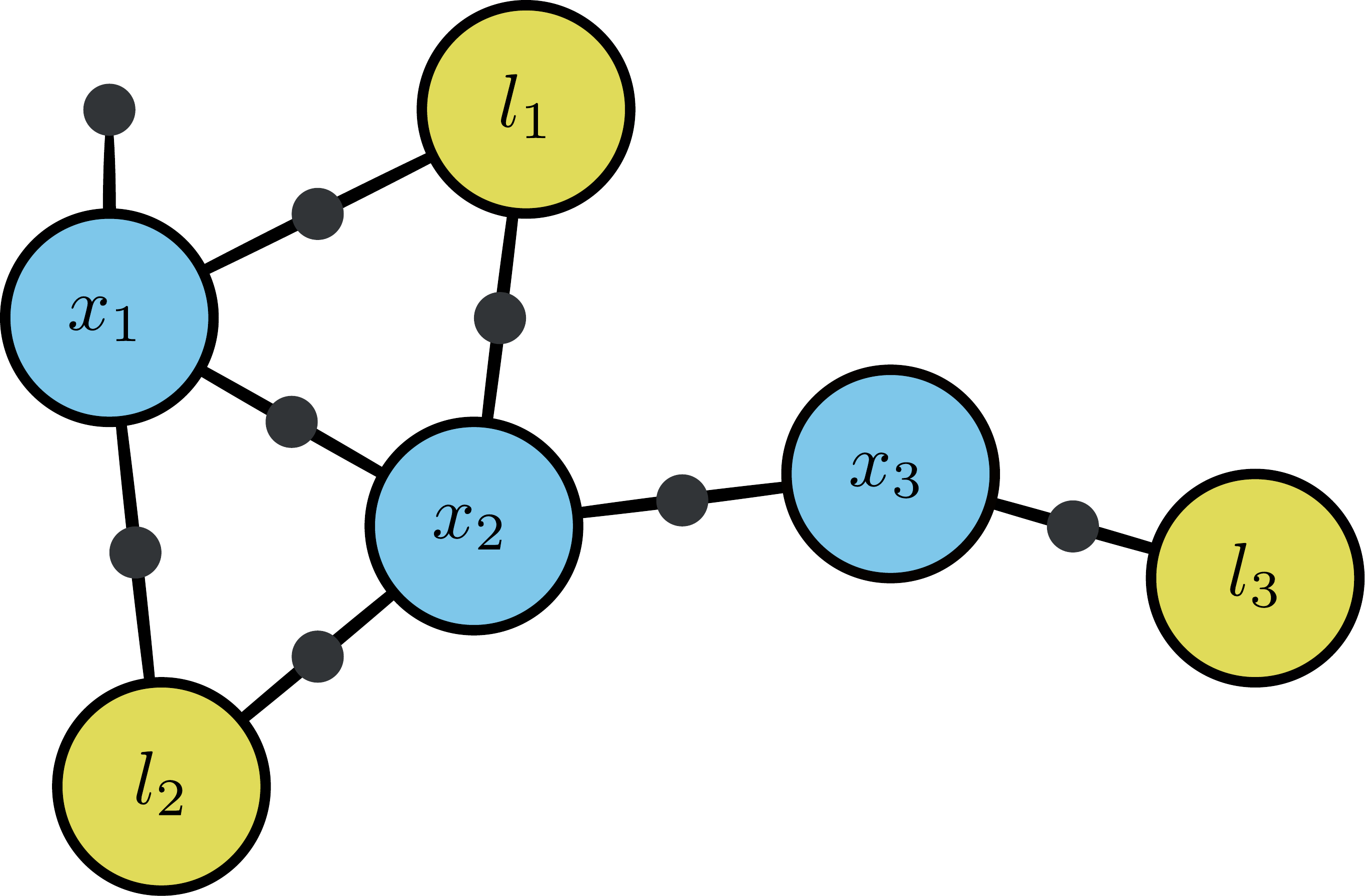}
\caption{An exemplary factor graph representing the belief of a SLAM agent. The state contains two types of variables: poses (blue nodes) and landmarks (yellow nodes). Motion and observation factors (small black nodes) connect the variables. }
\label{fig:example-factor-graph}
\end{figure}
 
This factorization also allows us to represent our system at each time-step with a common Probabilistic Graphical Model (PGM) -- the factor graph $FG_k \doteq (\mathcal{X}_k\cup\mathcal{F}_k,\mathcal{E}_k)$, where the set of nodes is comprised of the \emph{set} of state variables $\mathcal{X}_k$, and set of factors $\mathcal{F}_k$; and the undirected edges in $\mathcal{E}_k$ are used to connect state variables to the factors in which they are involved. An example is given in Fig.~\ref{fig:example-factor-graph}.

\subsection{\changedRemoved{Square-Root SAM: Inference with Linear/Gaussian Models \label{sec:prelim-square-root-sam}}}
A particularly interesting case, which we will focus on in this paper, is when the prior and model noise are Gaussian, i.e.,
\begin{eqnarray}
w_k &\sim& \mathcal{N}(0, \bm W_k),\\
v_k &\sim& \mathcal{N}(0, \bm V_k), \\
v_0 &\sim& \mathcal{N}(0, \bm V_0),
\end{eqnarray}
where $\forall k\in\mathbb{N},\, \bm W_k, \bm V_k$ are the covariance matrices of the respective zero-mean Gaussian noise variables $\bm w_k, \bm v_k$, and $\bm V_0$ is the covariance matrix of the prior over the pose $x_0$.

Assigning the corresponding Gaussian probability density functions in (\ref{eq:belief-factor-product}) yields 
\begin{multline}
\label{eq:belief-prod-exp}
b_k(\mathcal{X}_k) \propto \exp \left[ -\frac12 \norm{ x_0 - \overline{x_0} }^2_{\bm V_0}\right] \\\cdot \prod_{i=1}^{k} \exp \left[ -\frac12 \norm{x_i - g_k(x_{i-1}, u_i)}^2_{\bm W_i}\right] \\\cdot \prod_{i=1}^k \exp \left[ -\frac12 \norm{z_{k} - h_k(\mathcal{X}^o_k)}^2_{\bm V_i}\right],
\end{multline}
where the notation $\norm{\square}_{\bm\Sigma} \doteq \sqrt{\square^T{\bm\Sigma}^{-1} \square}$ marks the \mbox{Mahalanobis distance.}

\begin{figure}[t]
\includegraphics[width=\textwidth]{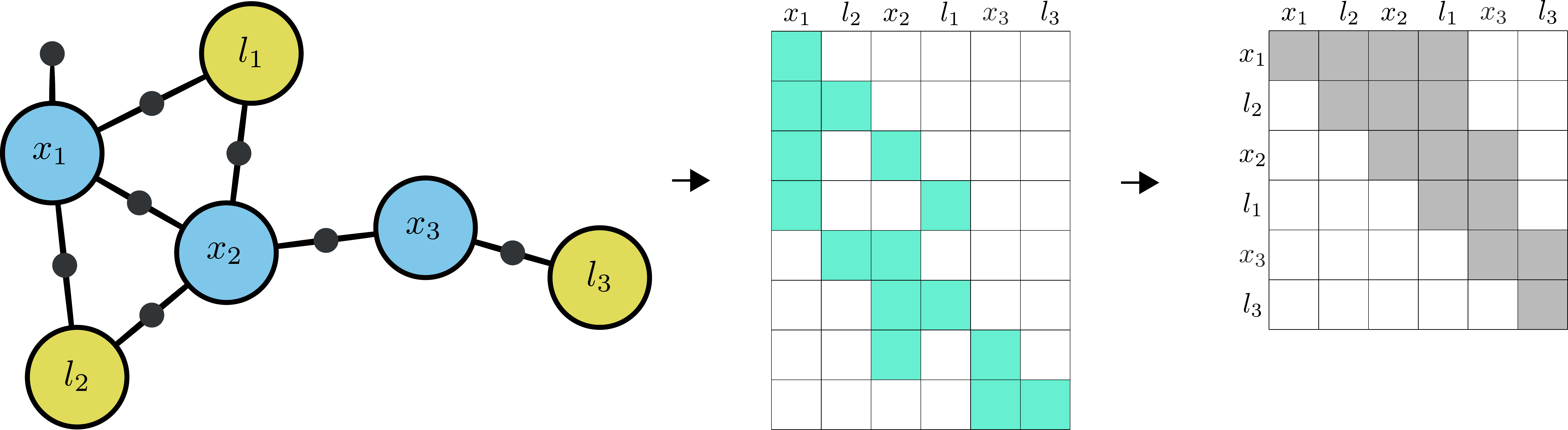}
\caption{When the models are linear(ized), the factor graph from Fig.~\ref{fig:example-factor-graph} (left) can be represented with the "Jacobian matrix" $\bm J$ (center). To perform inference, we shall perform QR factorization on $\bm J$, to derive the "square root" matrix $\bm R$ (right).}
\label{fig:example-qr-fact}
\end{figure}

Further, we may also consider the factor models are linear; otherwise, if the models are not linear, we can consider a local linearization of the non-linear functions, by calculating their Jacobian matrices, around an initial state estimate $\overline{\bm X_k}$ (typically, the MAP from time-step $k-1$, with proper initialization for new variables). Note that to calculate these Jacobians, the order of variables, which matches the order of columns in the matrices, must be fixed; we use \mbox{$\bm X_k \doteq [\bm X_k(1),\dots,\bm X_k(n_k)]^T$}, where $n_k$ is the state size, to mark the state \emph{vector}, which represents the order of state variables. Given these linearized models, we can easily derive that the belief at time-step $k$ (from (\ref{eq:belief-prod-exp})), is also (approximately) Gaussian:
\begin{equation}
b_k(\bm X_k) \approx \mathcal{N}({\bm  X}^*_k, \bm \Lambda_k^{-1}),
\end{equation}
where $\bm \Lambda_k$ is the information/precision matrix, achieved by calculating $\bm\Lambda_k \doteq \bm J_k^T \bm J_k$, where
\begin{equation}
\bm J_{k\geq1} \doteq \begin{bmatrix}
    \bm J_0 \\ \bm J^\delta_1 \\ \vdots \\ \bm J^\delta_k
\end{bmatrix}
,\, \bm J_0 \doteq \begin{bmatrix}
   \bm V_0^{-\frac12} \bm P_0
\end{bmatrix}
,\, \bm J^\delta_{k\geq1} \doteq \begin{bmatrix}
    \bm W_k^{-\frac12} \bm G_k \\ \bm V_k^{-\frac12} \bm H_k
\end{bmatrix},
\end{equation}
$g'_k(x_{k-1}, u_k, x_k) \doteq - x_k + g_k(x_{k-1}, u_k)$, 
$\bm G_k \doteq \nabla \vert_{\overline {\bm  X}_k} g'_k$, \linebreak
$\bm H_k \doteq \nabla \vert_{\overline {\bm  X}_k} h_k$, $\bm P_0 \doteq \nabla \vert_{\overline {\bm  X}_k} x_0$, and $\square^{\frac12}$ marks the Cholesky factor of the matrix. We note that since every row in the "collective" Jacobian matrix $\bm J_k$ represents a factor from $\mathcal{F}_k$, and the order of columns in the matrix matches the variable order in $\bm X_k$, the sparsity pattern of $\bm J_k$ matches the structure of the factor graph $FG_k$: looking at each row of the matrix, entries in it are non-zero, if and only if the variables matching the column-index of these entries are involved in the corresponding factor. This is demonstrated in Fig.~\ref{fig:example-qr-fact}.

In this case, the belief mean $\bm X_k^*$ matches the MAP estimate. It can be inferred as $\bm X_k^* = \overline{\bm X_k} + \bm\Delta^*_k$, where $\bm\Delta^*_k$ -- the "correction" to the initial estimate -- is achieved by solving the following linear "least squares" problem:
\begin{equation}
\bm J_k \cdot \bm\Delta_k  = \bm \zeta_k,
\end{equation}
with the Right Hand Side (RHS) vector $\bm\zeta_k$, given as
\begin{equation}
\bm\zeta_{k\geq1} \doteq \begin{bmatrix}
    \bm\zeta_0 \\ \bm\zeta^\delta_1 \\ \vdots \\ \bm\zeta^\delta_k
\end{bmatrix}
,\, \bm\zeta_0 \doteq
\begin{bmatrix}
   \overline{x_0}
\end{bmatrix}
,\, \bm\zeta^\delta_{k\geq1} \doteq 
\begin{bmatrix}
\overline{x_k} - g(\overline{x_{k-1}},u_k) \\
z_k - h(\overline{\mathcal{X}^o_k}) 
\end{bmatrix}.
\end{equation}

The solution of this system is achieved by calculating the upper triangular \emph{square root information matrix} $\bm R_k$ (in Fig.~\ref{fig:example-qr-fact}), given by the QR factorization of $\bm J_k$, or the Cholesky factorization of $\bm \Lambda_k$, and then performing "back substitution". From this square root matrix we can also conveniently infer other common properties of interest, such as marginal distributions, and the differential entropy. For more details, refer to \cite{Dellaert06ijrr}, or \cite{Elimelech21thesis}.

Non-zero entries in the square root matrix $\bm R_k$, which were zero entries in the information matrix $\bm \Lambda_k$, are known as "fill-in". The amount of fill-in is directly related to the computational cost of the factorization, and hence we care to minimize it. The amount of fill-in is determined by the variable (i.e., column elimination) order. While finding the optimal order is NP-complete (as mentioned), various heuristics -- variations of the minimum degree algorithm -- exist, and most prominently, $\mathtt{COLAMD}$ \citep{Davis04toms}. The amount of fill-in also determines the computational cost of performing back-substitution, though this cost is normally less significant than that of the factorization itself.

\subsection{\changedRemoved{Incremental Belief Update}}
According to the former explanation, incorporation of new factors to the belief can be seen as adding new linear constraints to the system. Hence, to update the Jacobian matrix from time-step~$k$, to time-step~$k+1$, after performing a control $u_{k+1}$, and receiving an observation $z_{k+1}$, we need to simply add new rows to it, such that
\begin{equation}
\bm J_{k+1} \doteq \begin{bmatrix}
   \begin{matrix} \bm J_k & \bm 0 \end{matrix} \\ -\, \bm J^\delta_{k+1} \,-
\end{bmatrix}.
\end{equation}
Note that in order for the columns in the "old" and "new" rows to match, this also requires proper augmentation of $\bm J_k$ with zero-columns, accounting for newly introduced variables $\mathcal{X}^\delta_{k+1}$. The placement of the augmented columns should be according to their index in the state vector $\bm X_{k+1}$; here we added the new variables at the end of the state. As mentioned, re-observing variables from the (non-immediate) past is commonly known as "loop closing".

\begin{figure}[t]
\includegraphics[width=\textwidth]{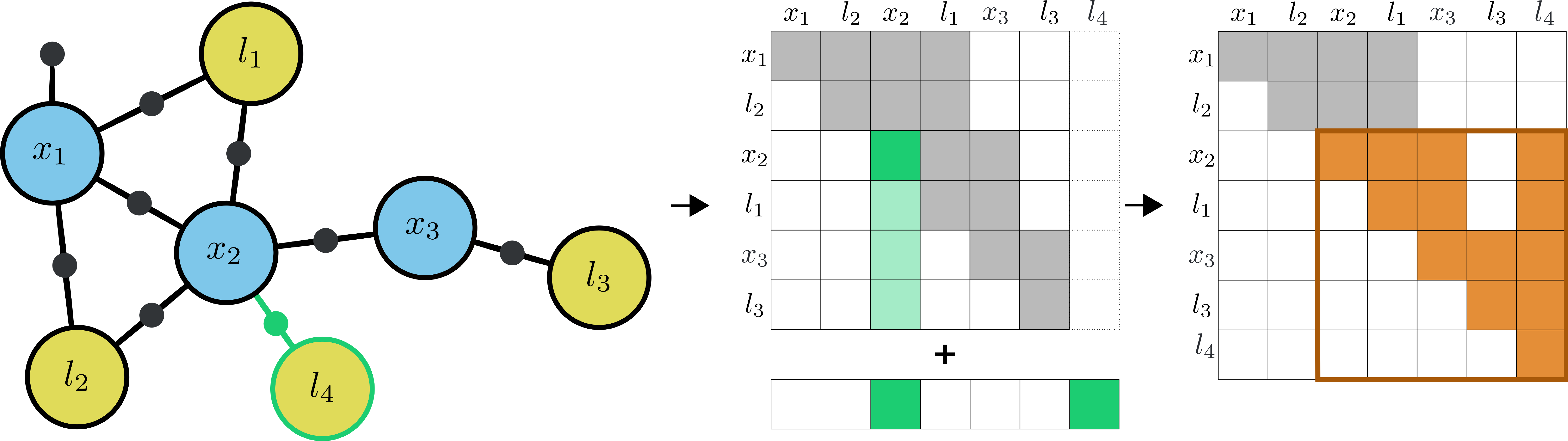}
\caption{New factors added to the system (left, in green), are represented with new Jacobian rows, with which we should update the square root matrix (center). The updated square root matrix (right) can be derived incrementally, by only recalculating the square bottom-right "affected" block, starting from the index of the first non-zero column in the update (in orange).}
\label{fig:inc-update}
\end{figure}

To update our belief and estimate, we shall calculate the "new" square root matrix $\bm R_{k+1}$. Fortunately, $\bm R_{k+1}$ can be derived by incrementally updating $\bm R_k$, instead of calculating the factorization of $\bm J_{k+1}$ (or $\bm \Lambda_{k+1}$) "from~scratch". First, we should identify the index of the first non-zero column in $\bm J^\delta_{k+1}$, and we mark it as $j$; this represents the index in the state vector of the first variable involved in the added factors. According to the QR algorithm, calculation of the top $j-1$ rows of $\bm R_k$ and $\bm R_{k+1}$ is equivalent, and thus we only need to calculate the bottom right block of $\bm R_{k+1}$, starting from the $j$-th row/column (the \emph{"affected block"}). This idea, which is visualized in Fig.~\ref{fig:inc-update}, is utilized by state-of-the-art "Smoothing and Mapping" (SAM) algorithms, such as iSAM1 \citep{Kaess08tro}, iSAM2 \citep{Kaess12ijrr}, and SLAM++ \citep{Ila17ijrr}. Of course, after the incremental re-factorization of the system, we shall again perform back sub substitution, if we wish to update the MAP estimate.

To take advantage of such incremental updates, the order of the "unaffected" first $j-1$ variables in $\bm X_{k+1}$ must match the one set in $\bm X_{k}$; i.e., the order of the first $j-1$ columns of $\bm J_{k+1}$ must match those of $\bm J_k$. It is possible, and sensible, to reorder columns $j$ to $n_{k+1}$ during the re-factorization; e.g., to apply a fill-reducing order. Such incremental application of fill-reducing ordering tactics is likely to be inferior, in terms of the resulting fill-in, than if we were to consider reordering of all the variables, but proves to still be effective. To clarify, it is not possible to reorder the variables in order to make the affected block (which is due to re-factorization) smaller.

\subsection{\changedRemoved{Belief Space Planning \label{sec:prelim-planning}}}
Up until now, we examined a "passive" problem, in which we cared to estimate an agent's state, given known actions and observations. Understanding this problem was essential for describing the planning problem, to its solution we wish to contribute with this work.

When planning, we wish to actively seek the next course of action for the agent; if we consider a SLAM agent, as we have done so far, this problem is known as "active~SLAM". Thus, assume the agent currently holds the belief $b_k$ over its state~$\mathcal{X}_k$. The goal is to select (online) the optimal policy \mbox{$\pi\in\Pi$} -- a mapping from beliefs to actions -- which, when applied $T$ times ("the planning horizon"), starting from~$b_k$, would maximize the expected accumulated discounted reward ("the expected return"), as measured with the value or objective function:
\begin{multline}
\label{eq:value-function}
V\left(b_k, \pi\right) \doteq \\ \expect_{\mathcal{Z}_{k+1:k+T-1}} \left[ \sum_{t=1}^T \gamma_t\cdot\left[\rho\left(b_{k+t-1}, u_{k+t}\right) \mid \pi,\mathcal{Z}_{k+1:k+t-1}\right]\right],
\end{multline}
where, for $t\geq1$, $\mathcal{Z}_{k+1:k+t-1} \doteq \left\{z_{k+1},\dots,z_{k+t-1}\right\}$; $u_{k+t} = \pi(b_{k+t-1})$; and the belief $b_{k+t-1}$ is recursively defined as in (\ref{eq:belief-factor-product}). Note that for generality, we mark the discount factor $\gamma_t\in[0,1]$ as a function of the index, to not force reward diminishment.

Since our objective and policies are belief-dependent, this problem is known as planning in the "belief space". Specifically, in information-theoretic belief space planning (BSP), such as in active SLAM, we often wish to minimize the uncertainty in the posterior belief. We can use the differential entropy as an uncertainty measure, which, for a Gaussian belief $b$, over a state of size $n$, with an information matrix $\bm \Lambda = \bm R^T \bm R$, is
\begin{equation}
\entropy\left(b\right) = \frac12\cdot\ln\left[\frac{\left(2\pi e\right)^{n}}{\abs{\bm \Lambda}}\right] = -\left( \ln \abs{\bm R} - \frac n 2 \cdot\ln \left(2\pi e\right) \right).
\end{equation}
Then, we can define the following objective function:
\begin{equation}
\label{eq:objective}
\tilde{V}\left(b_k, \pi\right) \doteq \expect_{\mathcal{Z}_{k+1:k+T}} \left[ \entropy(b_{k})-\entropy(b_{k+T}\mid \pi,\mathcal{Z}_{k+1:k+T}) \right],
\end{equation}
which measures the expected information gain between the current and final beliefs. As we can see, this value depends on the square root matrix of the agent's belief at the end of the candidate trajectory.

\subsection{\changed{Problem Definition} \label{sec:prelim-def}}
Overall, we can formulate the following decision problem: given an initial belief $b_k$ over the state $\mathcal{X}_k$, with known transition and observation models, a set $\Pi$ of candidate policies, and a value function $V$, such as the one defined in~(\ref{eq:objective}), return the optimal candidate $\pi^*$, such that
\begin{equation}
\label{eq:decision-problem}
\pi^* = \operatorname*{argmax}_{\pi \in \Pi} V\left(b_k,\pi\right).
\end{equation}

Since solving this problem requires to perform belief updates considering multiple candidate policies, and over long horizons, the computational cost of the solution can turn exceptionally high. The goal of this work is to help \emph{alleviate the computational cost} of this problem's solution. Note that we may also consider other value functions, as this choice is mostly orthogonal to the contribution to follow.

\subsubsection{Assumptions and Generalizations}
Evidently, to calculate the candidates' expected returns, the prior belief should be predictively propagated (updated) according to each of the candidate policies. Yet, since our models are stochastic, utilization of each policy may lead to different outcomes, or \emph{hypotheses}; in our active SLAM formulation, each hypothesis can be described as an assignment to the Markov chain $H_{k+1:k+T} \doteq [u_{k+1},z_{k+1},\dots,u_{k+T},z_{k+T}]$, according to the candidate policy, and the observation model. Luckily, it was shown that using only \emph{a single hypothesis per candidate} is often enough to achieve sufficient accuracy, by assuming "Maximum Likelihood (ML)" observations (see \cite{Platt10rss}); it is hence a common assumption taken to practically solve this online planning problem. For conciseness of the discussion, we will utilize this assumption when describing our approach in the following sections; however, this is not an inherent limitation of the approach. An extended discussion on observation prediction and generalization of the approach, to allow multiple hypotheses per candidate, is provided in Appendix~\ref{sec:appendix-general-obs}.

For the same reason, we will also assume linear(ized) and Gaussian models, which means our belief can be described and updated using the aforementioned matrix-based formulation; yet again, this is not an inherent limitation of the approach, which is applicable to more general models as well. An extended discussion generalizing the approach to other models, using PGMs, is provided in Appendix~\ref{sec:appendix-general-models}.

%% file: approach.tex
\section{PIVOT: Predictive Incremental Variable Ordering Tactic \label{sec:pivot}}
\subsection{\changed{Overview}}
\begin{figure}[t]
		\includegraphics[width=0.68\textwidth]{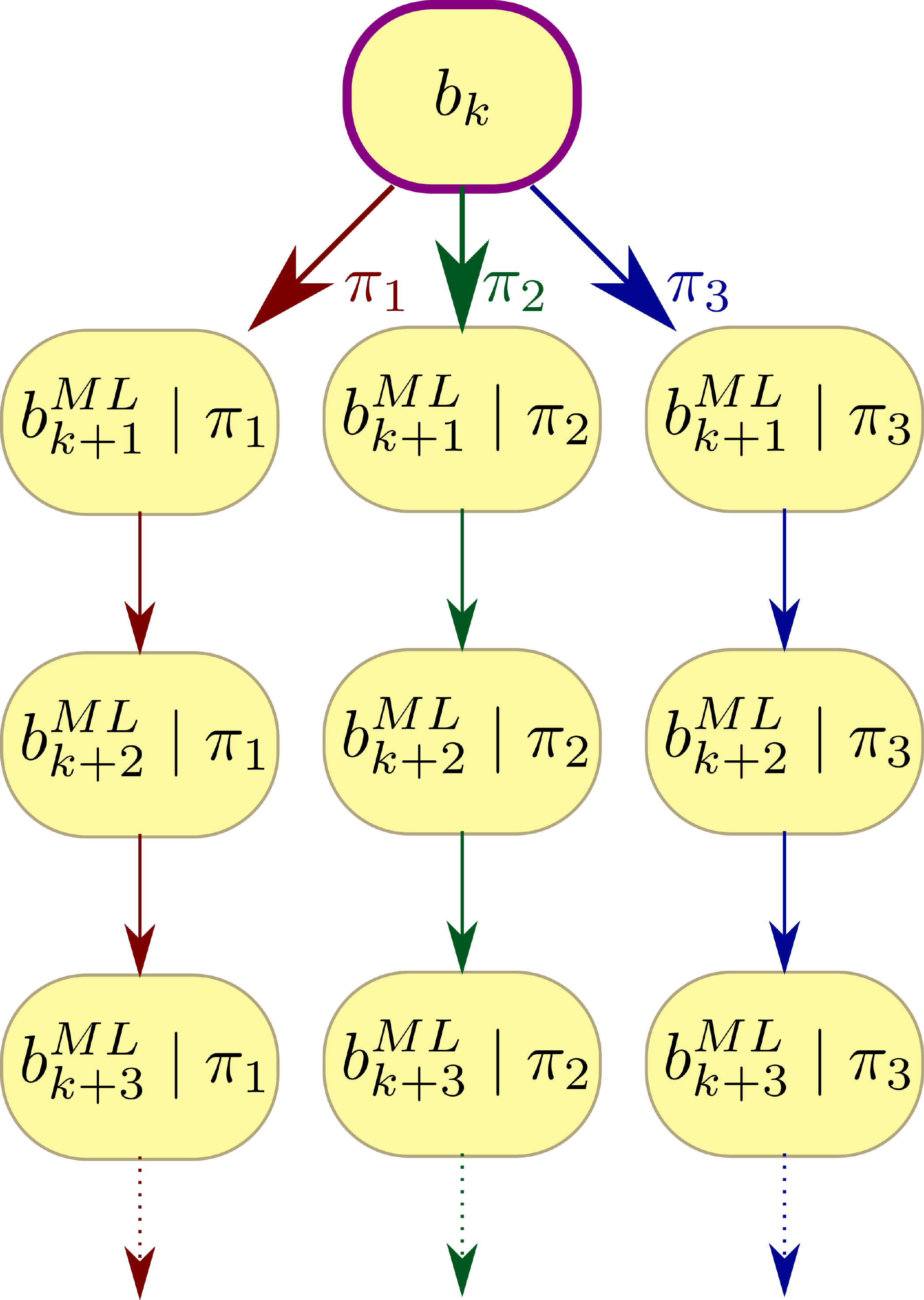}
        \caption{A planning tree, connecting predicted (ML) hypotheses for three candidate policies at their root, which is the initial belief (in purple). To benefit all future updates, we shall apply $\pivot$ once on the initial belief.}
		\label{fig:planning-tree-ml}
\end{figure}

Say we are interested in performing a planning session, as described before. At this point in time, to calculate the candidate policies' values, we should perform multiple updates to the current belief "in parallel", propagating it according to the predicted hypotheses of the candidates, as demonstrated in Fig.~\ref{fig:planning-tree-ml}. We thus suggest the following concept: optimize the initial belief representation predictively -- i.e., before and in preparation for these updates -- in order to increase their efficiency. In practice, we wish to perform a single, precursory, and standalone modification to the initial belief's square root matrix, to convey \emph{reordering of the state variables}. We call this idea $\pivot$: {Predictive Incremental Variable Ordering Tactic.} Although applying $\pivot$, which may require partial matrix re-factorization, comes at a certain cost, this small investment in the current time should be overall beneficial in the long run, by reducing the cost of all future belief updates. It is important to mention that this modification simply conveys a change of representation; it is, hence, does not affect the accuracy of the belief, and does not compromise the quality of solution.

In the following sub-sections, we introduce the basic variation of $\pivot$ (in Section~\ref{sec:pivot-push}), and several extensions to it, which can lead to even greater improvement in performance: multi-class (Section~\ref{sec:pivot-multi}), fill-aware (Section~\ref{sec:pivot-fill}), and forced-incremental (Section~\ref{sec:pivot-inc}).

\begin{figure}[ht]
\includegraphics[width=\textwidth]{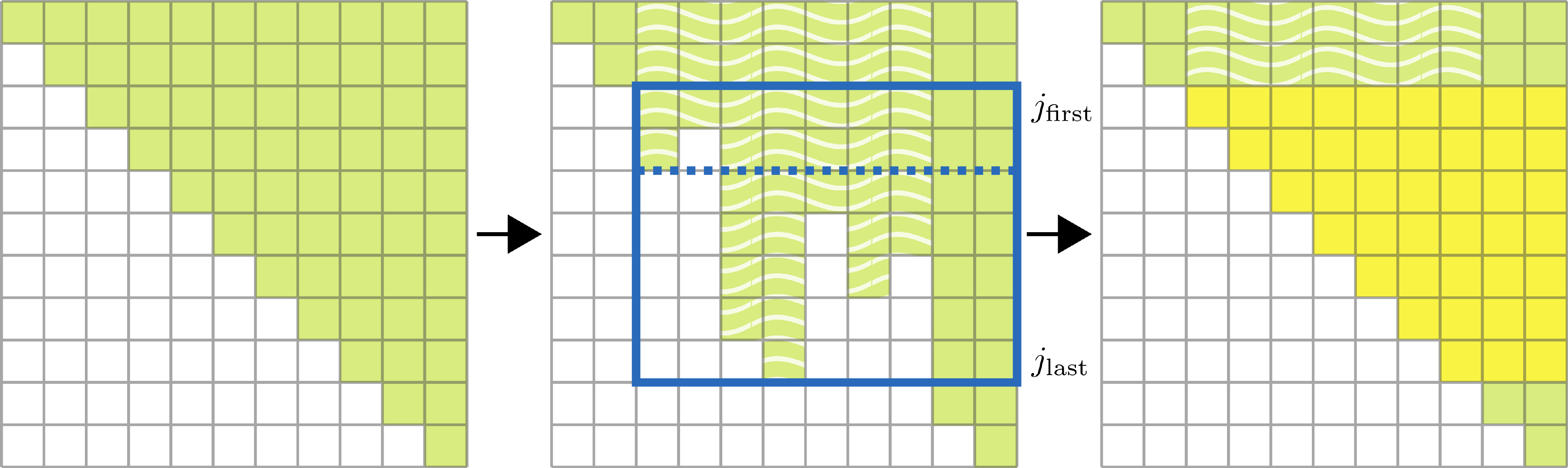}
\caption{The state variable order matches the column order in the square root matrix $\bm R$ (on the left). Variable reordering hence implies reordering of the columns (white overlay), but this would break the matrix' upper triangular structure. Thus, to apply a new variable order, the rows indexed between the first and last permuted indices (boxed in blue), should be recalculated.}
\label{fig:reordering}
\end{figure}

\subsection{Determining vs. Applying a New Order \label{sec:applying-order}}
To avoid confusion, before describing $\pivot$, we wish to clarify that discussion to follow only comes to answer the question "what should be the new variable order?". The question of "how to modify the existing square root matrix $\bm R$, after the new order is determined?" was extensively discussed in our previous work \citep{Elimelech21ral}, and one should refer to it for full details.

For good measure, we shall briefly mention some of the key points identified. First, we recall that the variable order matches the order of columns in $\bm R$. However, trying to reorder the variables by simply reordering the columns of the matrix, would break its upper triangular shape. Thus, when reordering the variables, we must re-calculate (a subset of) the rows indexed between $j_\text{first}$, and $j_\text{last}$, where those mark the index of the first and last permuted variables; a visualization of this idea is brought in Fig.~\ref{fig:reordering}. These affected rows can be recalculated through direct modification of $\bm R$, with no access to $\bm J$ nor the original factors; yet, if we also wish to recalculate the cache of marginal factors (which is needed for future updates), we must calculate these rows through partial re-factorization. Again, the full details of this process are covered in the aforementioned work.

\subsection{\changed{Pushing Forwards Involved Variables \label{sec:pivot-push}}}
Let $b(\bm X)$ mark the initial belief, and $FG \doteq (\mathcal{X}\cup\mathcal{F},\mathcal{E})$ its corresponding factor graph; in the linear/Gaussian~case, this belief is represented with the mean vector $\overline{\bm X}$, and information matrix $\bm \Lambda \doteq \bm J^T \bm J$, such that $b \approx \mathcal{N}(\overline{\bm X}, \bm \Lambda)$. 
Also, let $\mathcal{H} \doteq \left\{ H_1,\dots,H_m \right\}$ represent the set of candidates' hypotheses, which, in our case, are of the form $H\doteq[u_1,z_1,\dots,u_t,z_t]$.

\pagebreak

\begin{figure}[t]
\includegraphics[width=0.72\textwidth]{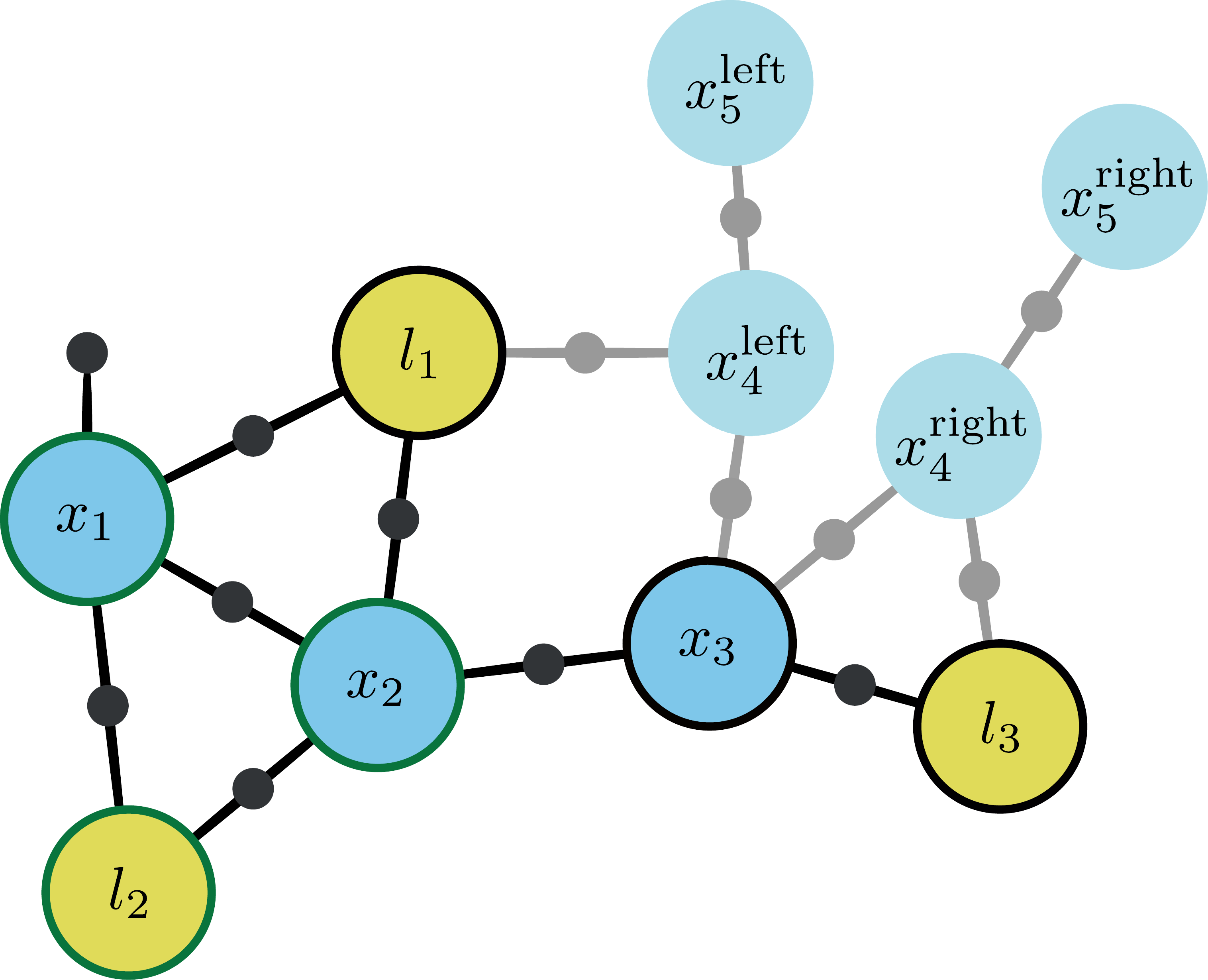}
\caption{Identifying involved variables. The factor graph from Fig.~\ref{fig:example-factor-graph} represents the initial belief, which is extended according to two candidate policies: following the right and left paths. We hypothesize that each path would lead to the addition of two new poses, and three new factors (in gray). For each hypothesis we can identify the involved variables (from the initial state), as those connected to the new factors: $\inv_\mathcal{X}(\text{right}) = \{ x_3, l_3 \}$ and $\inv_\mathcal{X}(\text{left}) = \{ x_3, l_1 \}$; $\ninv_\mathcal{X}(\{ \text{right, left} \}) = \{ x_1, l_2, x_2 \}$ are uninvolved in any of the hypotheses. For planning, we are interested in updating the initial belief's square root matrix (which appears in Fig.~\ref{fig:example-qr-fact}), according to each hypothesis.}
\label{fig:supergraph}
\end{figure}

Each hypothesis~$H$ can be matched with its own factor graph $FG^\delta_H \doteq (\mathcal{X}\cup\mathcal{X}^\delta_H\cup\mathcal{F}^\delta_H,\mathcal{E}^\delta_H)$ ("update"), containing the variables and constraints we wish to integrate into the initial belief, and with which we wish to extend the initial factor graph, such that $ FG_H \doteq FG \cup FG^\delta_H $. By looking at $FG^\delta_H$, for each hypothesis $H\in\mathcal{H}$, we can identify the set of involved prior-state variables marked as $\inv_\mathcal{X}(H)$: these are the variables which are directly involved in any of the "new" factors $\mathcal{F}^\delta_H$. The remaining variables are uninvolved in this update, and marked $\ninv_\mathcal{X}(H)$. This is visualized in Fig.~\ref{fig:supergraph}. Accordingly, the variables which are involved in at least one update $H\in\mathcal{H}$ are marked $\inv_\mathcal{X}(\mathcal{H})$, and the variables which are uninvolved in any of the updates are marked $\ninv_\mathcal{X}(\mathcal{H})$. We use the subscript $\bm X$, instead of~$\mathcal{X}$, to refer to the respective subset of variables, \emph{ordered} according to the variables' index in $\bm X$. In the linear/Gaussian case, each update~$H$ is matched with a matrix $\bm J^\delta_H$ containing the new rows with which we want to extend the matrix $\bm J$; then, the involved variables can also be identified simply by the non-zero columns in the matrix $\bm J^\delta_H$. 

We now wish to update $b$ according to each of these hypotheses. We recall that to do so, we are actually interested in updating the belief factorization, which is represented with the square root information matrix~$\bm R$. We previously explained that (1) $\bm R$ depends on the initial variable order expressed in $\bm X$; and (2) updates can be performed via incremental re-factorization -- only considering the block of "affected variables" (i.e., all the variables starting from the first variable involved in the update). 
Notably, the initial variable order determines the index of the first involved variable in each of the future updates, and, by such, which variables would be affected by each of them. We should consider that updating the square root factorization has, at worst, cubical complexity in relation to the number of affected variables \citep{Hammerlin12book}; hence, for computational efficiency concerns, we wish this number, in each update, to be minimal.

\pagebreak

\begin{figure}[ht]
	\begin{subfigure}[t]{\textwidth}\center
		\includegraphics[width=\textwidth]{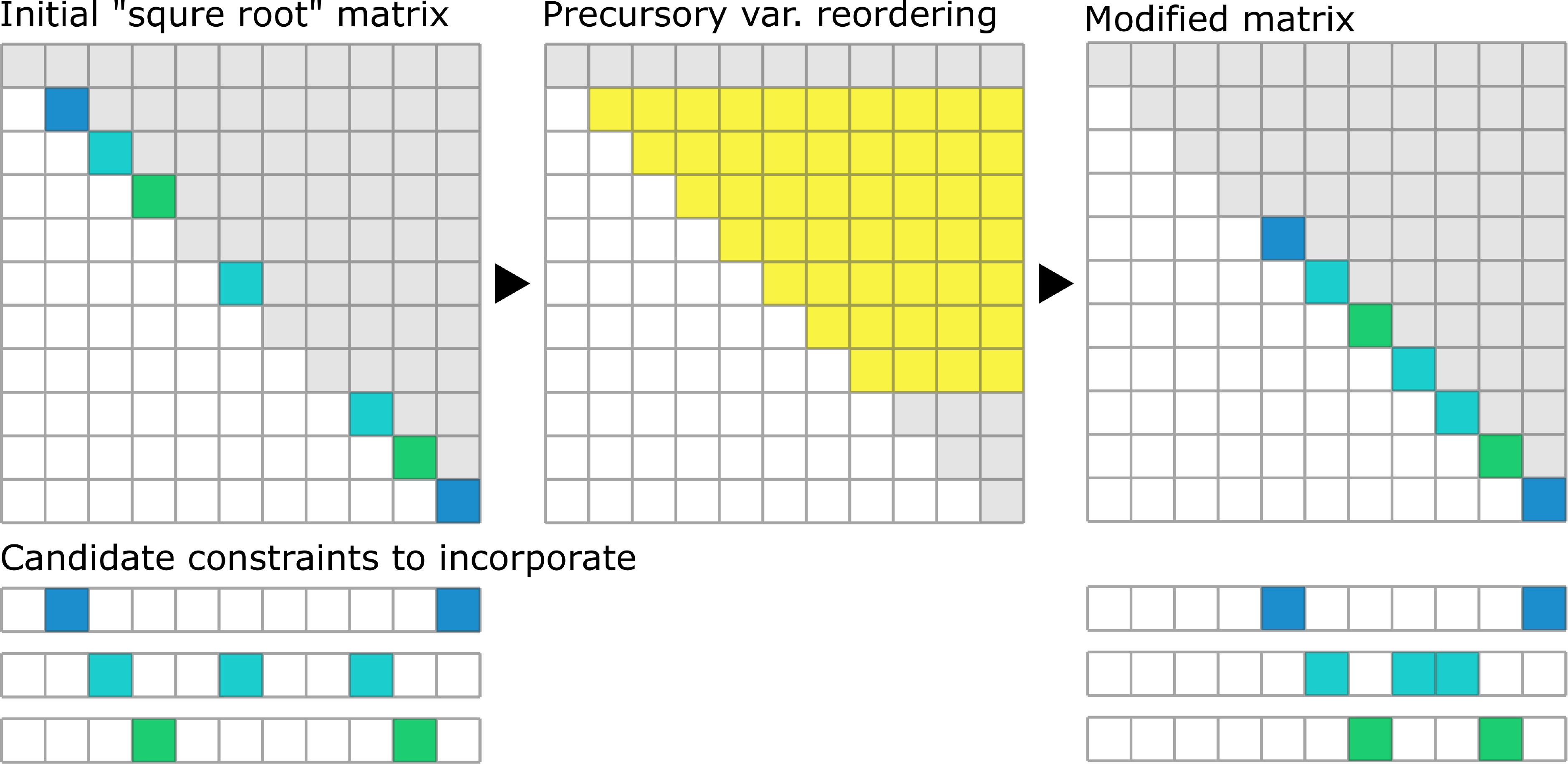}
        \caption{Planning with and without applying $\pivot_1$. On the left are the sparsity patterns of the square root matrix $\bm R$ of the initial belief $b$, and the Jacobian matrices of three candidate updates. 
        The order of columns in the matrices corresponds to the state variable order. 
        Colored cells in each update's Jacobian matrix signify non-zero columns, which in turn represent the involved variables in the update. According to $\pivot_1$, we should reorder the state variables, and push forwards the involved variables, towards the end of the state. The modified matrices, after variable reordering, appears on the right. Note that to keep its triangular shape, $\bm R$ requires specialized modification, as explained in Section~\ref{sec:applying-order}.
\\ $ $}
	\end{subfigure}
	
	\begin{subfigure}[t]{\textwidth}\center
		\includegraphics[width=\textwidth]{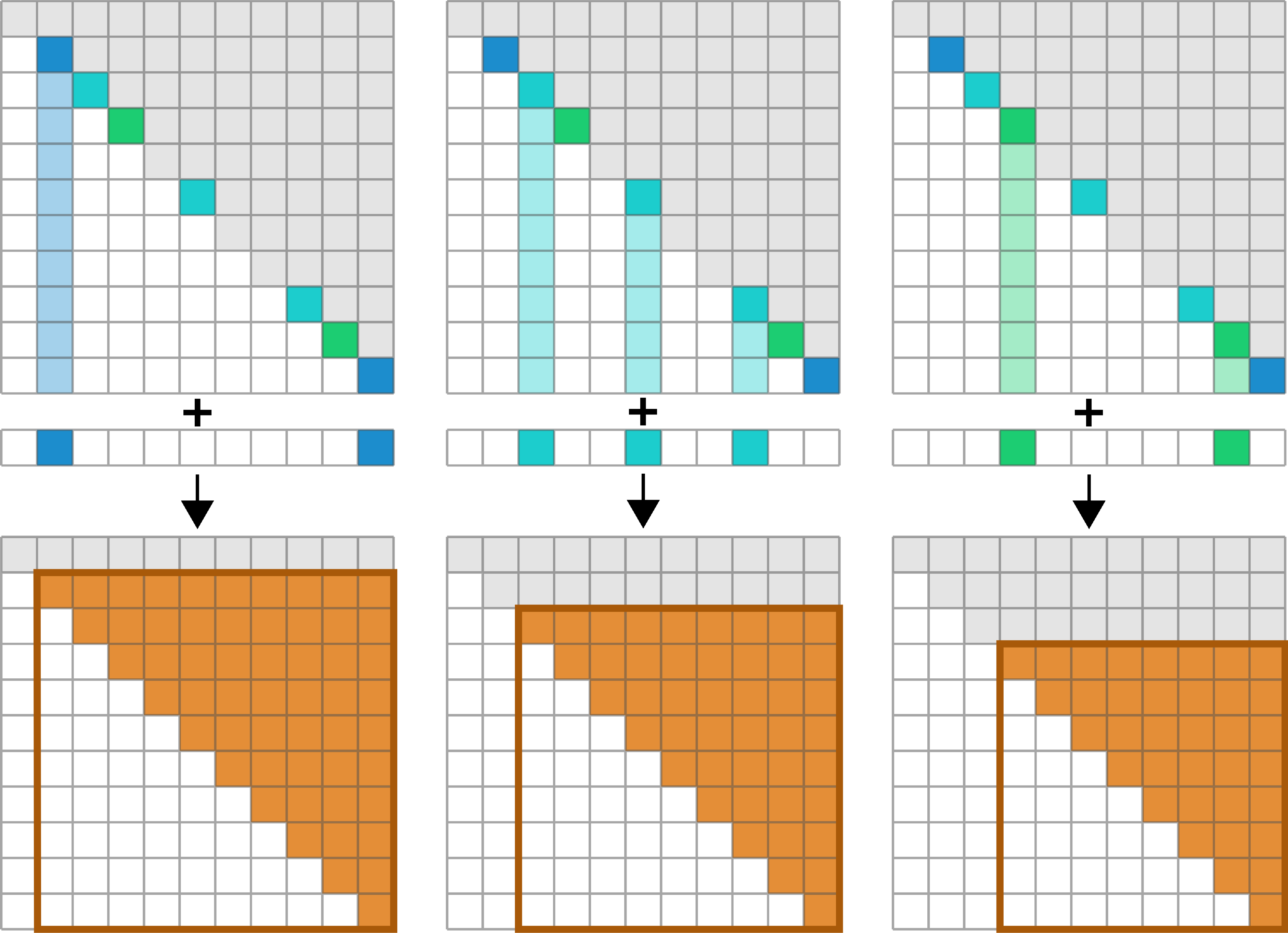}
        \caption{Updating the original matrix according to each of the candidates. The orange blocks in each posterior matrix represent the affected (i.e., recalculated) entries; this block is determined according to the index of the first involved variable (column).\\ $ $}
	\end{subfigure}
	
	\begin{subfigure}[t]{\textwidth}\center
		\includegraphics[width=\textwidth]{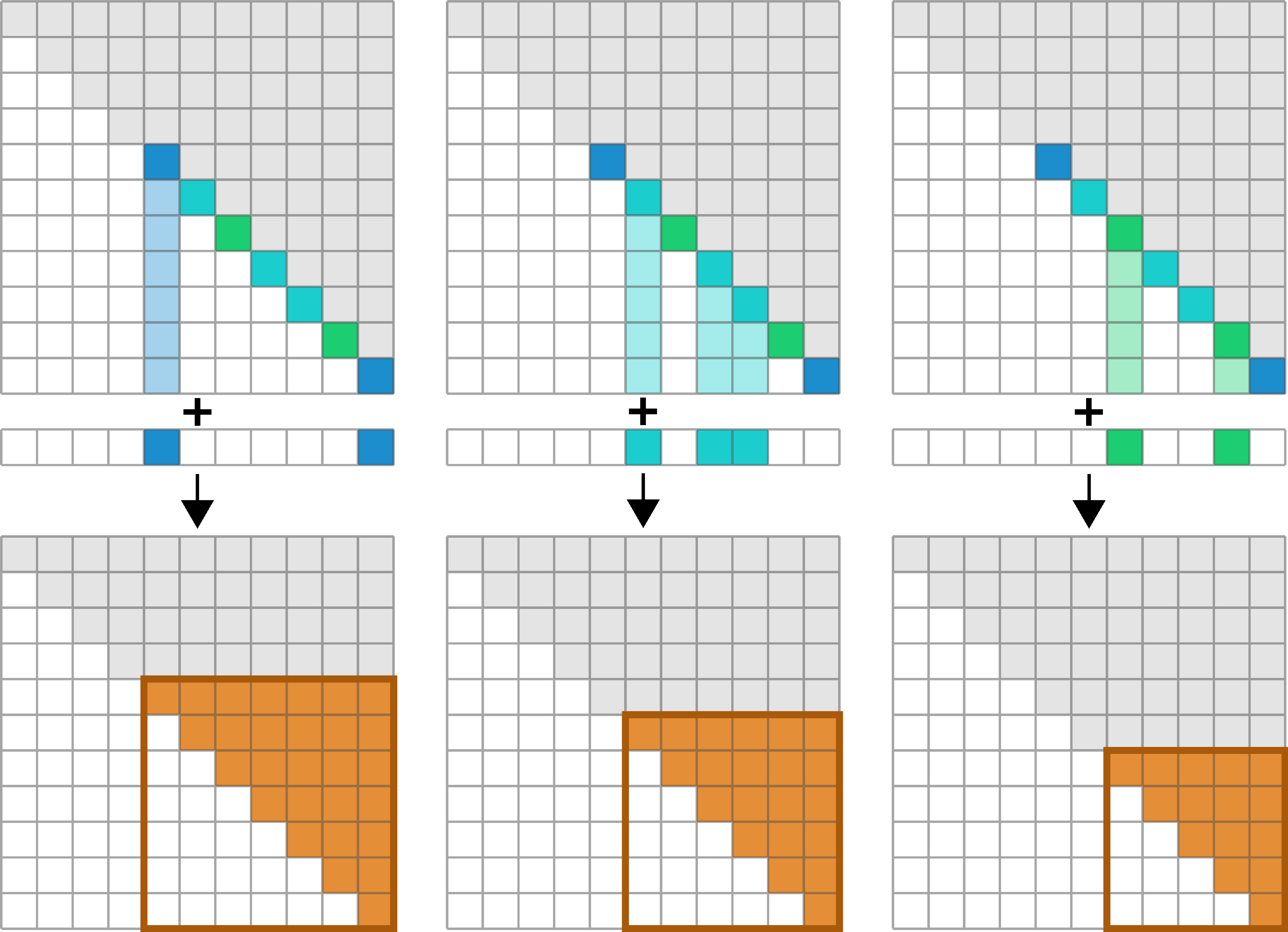}
        \caption{By adding a $\pivot_1$ reordering step before planning, we are able to reduce the size of the affected block in each candidate update.}
	\end{subfigure}

\caption{A conceptual demonstration of $\pivot_1$. Please note that we do not assume dense matrices; gray or orange-colored cells only indicate entries which \emph{may} be non-zero.}
\label{fig:pivot-concept}
\end{figure}

We note that for each update, the set of affected variables may include also uninvolved variables, which appear after the first involved variable (according to the initial variable order); such variables are only affected due to their index in the state. We will develop our variable ordering tactic to avoid such scenarios.
Beforehand, let us clarify our terminology, as used throughout this paper: pushing a variable forwards means increasing its index in the state vector, bringing it closer to the vector's end; accordingly, pushing a variable backwards means decreasing its index, bringing it closer to the start.

Thus, according to our ordering tactic, we suggest to push backwards all the uninvolved variables $\ninv_\mathcal{X}(\mathcal{H})$, to appear before $\inv_\mathcal{X}(\mathcal{H})$, or equally, \textbf{push the involved variables forwards}; this verifies that variables that are uninvolved in all the actions, are never affected while planning. This approach, which we refer to as \emph{$\pivot$: Predictive Incremental Variable Ordering Tactic}, can be represented with the permutation operator
\begin{equation}
\pivot_1(\bm X) \doteq \begin{bmatrix}
\ninv_{\bm X}(\mathcal{H}) \\ \inv_{\bm X}(\mathcal{H})
\end{bmatrix},
\end{equation}
which separates the variables according to their \emph{class}: involved or uninvolved. Optimally, for each hypothesis, we would like to update only the involved variables, and make all the uninvolved variables unaffected. However, since we perform the reordering only \emph{once per planning session}, some of the affected variables might still not actually be involved, for each candidate action. Nonetheless, such reordering maximizes the number of unaffected variables, and, by such, the overlapping sub-matrix of the prior square root matrix, which can best be reused when calculating all posteriors. This concept is demonstrated in Fig.~\ref{fig:pivot-concept}.

\subsection{Multi-Class PIVOT \label{sec:pivot-multi}}
We note that the previous separation of variables maintains the \emph{relative} order of variables in each class; i.e., if a variable $x$ initially came before variable $y$, and they both are in the same class, then even after the reordering, their relative order is kept. This means that involved variables may only be pushed forwards (compared to their original index), and the number of the affected variables by each planning hypothesis may only decrease.

Nonetheless, while maintaining the relative variable order in the "involved" class presumably makes sure that the cost of each update it reduced, this property is not always desirable when considering the overall planning performance point of view. For example, if a variable is only involved in one hypothesis, it might not be sensible to keep it ahead of another variable which is involved in all actions, as it would cause the first variable to unnecessarily be affected in all updates. Instead, we might want \textbf{the order among the involved variables to convey their "involvement level"}, i.e., the number of hypotheses in which each variable is involved. In this case, pushing "more forwards" variables which are "more involved", may increase the number of affected variables in some of the updates, but the total number of affected variables, considering all the updates, would be minimized.

\pagebreak

\begin{figure}[t]
\includegraphics[width=0.9\textwidth]{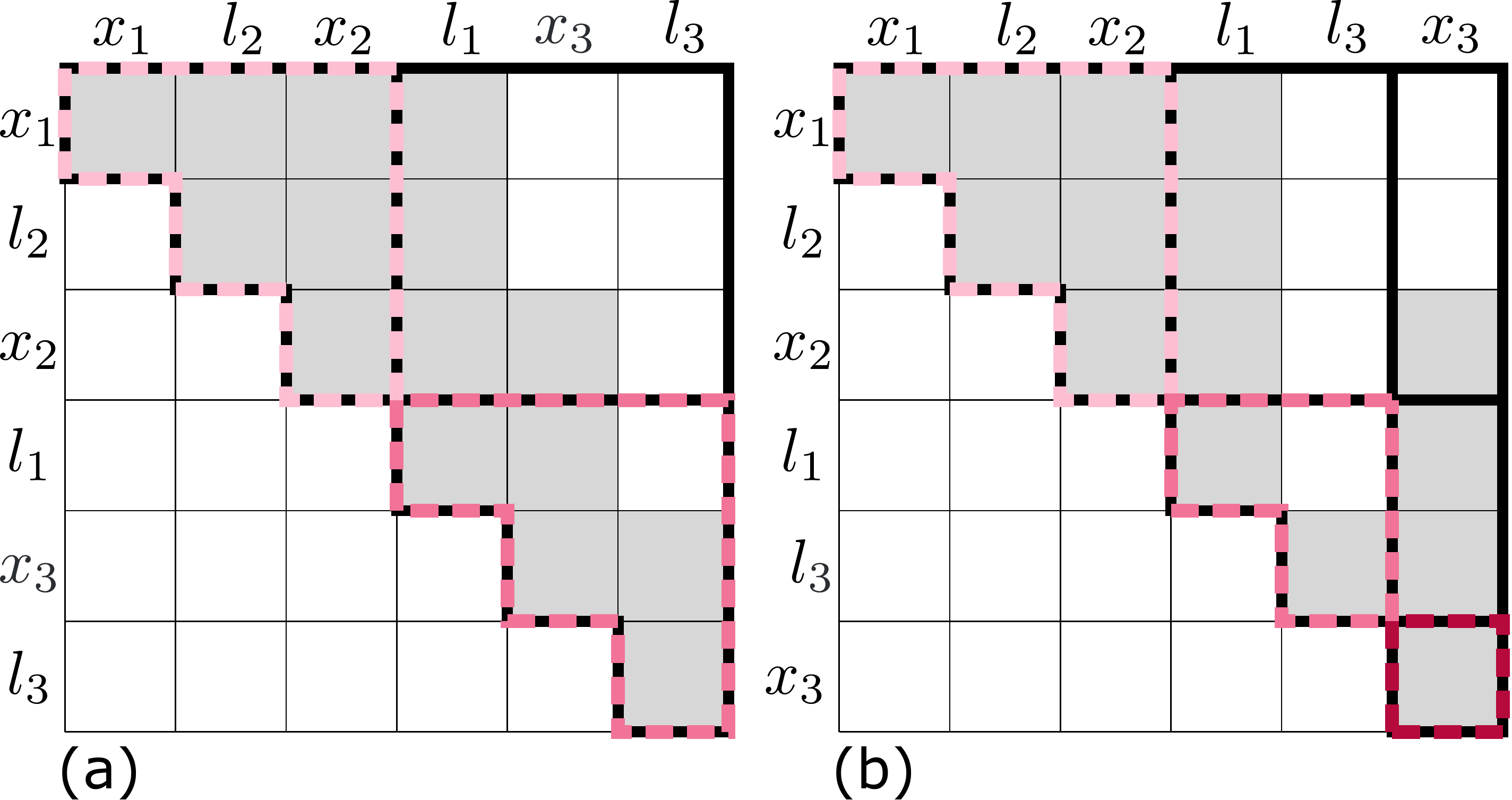}
\caption{The block structure of the modified square root matrix $\bm R$ of the prior belief from Fig.~\ref{fig:supergraph}, after applying $\pivot_1$ (a) and $\pivot_2$ (b). $\pivot_1$ puts all involved variables ($l_1,x_3,l_3$) in the same class; $\pivot_2$ separates the involved variables into two classes, as $x_3$ is involved in both hypotheses, while $l_1$ and $l_3$ are involved in one hypothesis, each.}
\label{fig:root-blocks}
\end{figure}

We thus suggest a generalization of $\pivot_1$, in which we subdivide the involved variables into $c\in \mathbb{N}$ (sub-)classes, according to their involvement level:
\begin{equation}
\pivot_c(\bm X) \doteq \begin{bmatrix}
\text{class}^0_{\bm X} \\ \vdots \\ \text{class}^c_{\bm X}
\end{bmatrix},
\end{equation}
where
\begin{align}
\text{class}^i_\mathcal{X} &\doteq \left\{ x\in\mathcal{X} \mid (i-1)\cdot\frac{M}{c} < \mathtt{level}(x) \leq i\cdot\frac{M}{c} \right\} \\
M &\doteq \max_{x\in \mathcal{X}} \mathtt{level}(x) \\
\mathtt{level}(x) &\doteq \sum_{H\in\mathcal{H}} 1\left(x\in\inv_\mathcal{X}(H)\right)
\end{align}
For example, when $c=1$, the class separation is based on a binary indicator of involvement -- this matches the definition of $\pivot_1$ from before, where all the involved variables are in the same class. When $c=2$, we subdivide the involved variables into two classes -- those which are involved in at most $\frac{M}{2}$ updates, and those who are involved in more than that amount (where $M$ is the maximum number of updates in which any variable is involved). 
We can similarly continue to increase $c$ for a finer division. We may also explicitly require $c=\max$, to treat every involvement level as a distinct class.
For any choice of $c$, $\text{class}^0_\mathcal{X} \equiv \ninv_\mathcal{X}(\mathcal{H})$ always contains the uninvolved variables. Algorithm~\ref{alg:pivot} summarizes the steps for calculating $\pivot_c$.
Further, each chosen $c$ imposes a block-diagonal structure on the modified square root matrix, separating the variables according to their class division, as demonstrated in Fig.~\ref{fig:root-blocks}.

\subsection{\changed{Fill-Aware PIVOT \label{sec:pivot-fill}}}
As reviewed, pushing the involved variables forwards is an ordering tactic \emph{intended} to keep the cost of future belief updates during planning low; however, this tactic is not always optimal. 
Indeed, when handling dense matrices (beliefs), then the cost of re-factorization (updates) is determined solely by the number of variables in the affected blocks. Yet, in sparse systems, the cost of such updates also tightly depends on the density of these blocks.

It is theoretically possible that applying the previous tactic, and reducing the size of an affected block, would actually make the cost of re-factorization higher, if this operation introduced more fill-in in this block (i.e., made it denser). Of course, we refer here to the density of the affected block in the \emph{prior} square root matrix; yet, this acts as the baseline for the fill-in in the re-calculated block in the posterior matrix, which can only increase after adding new constraints. In other words, a dense (block in the) prior would lead to a dense (block in the) posterior, which conveys high re-factorization cost. This fill-in would also affect the cost of performing back-substitution, which is needed in order to update the MAP estimate. Further, the fill-in in the prior also conveys the cost of the reordering process itself, which also typically requires re-factorization. Thus, we would like to optimize the variable order such that we find \textbf{\mbox{balance between reducing} the size of the affected blocks, and keeping the system density low}.
To achieve that, we introduce next two fill-aware improvements to the basic $\pivot_c$ order.

\begin{figure}[b]
\includegraphics[width=\textwidth]{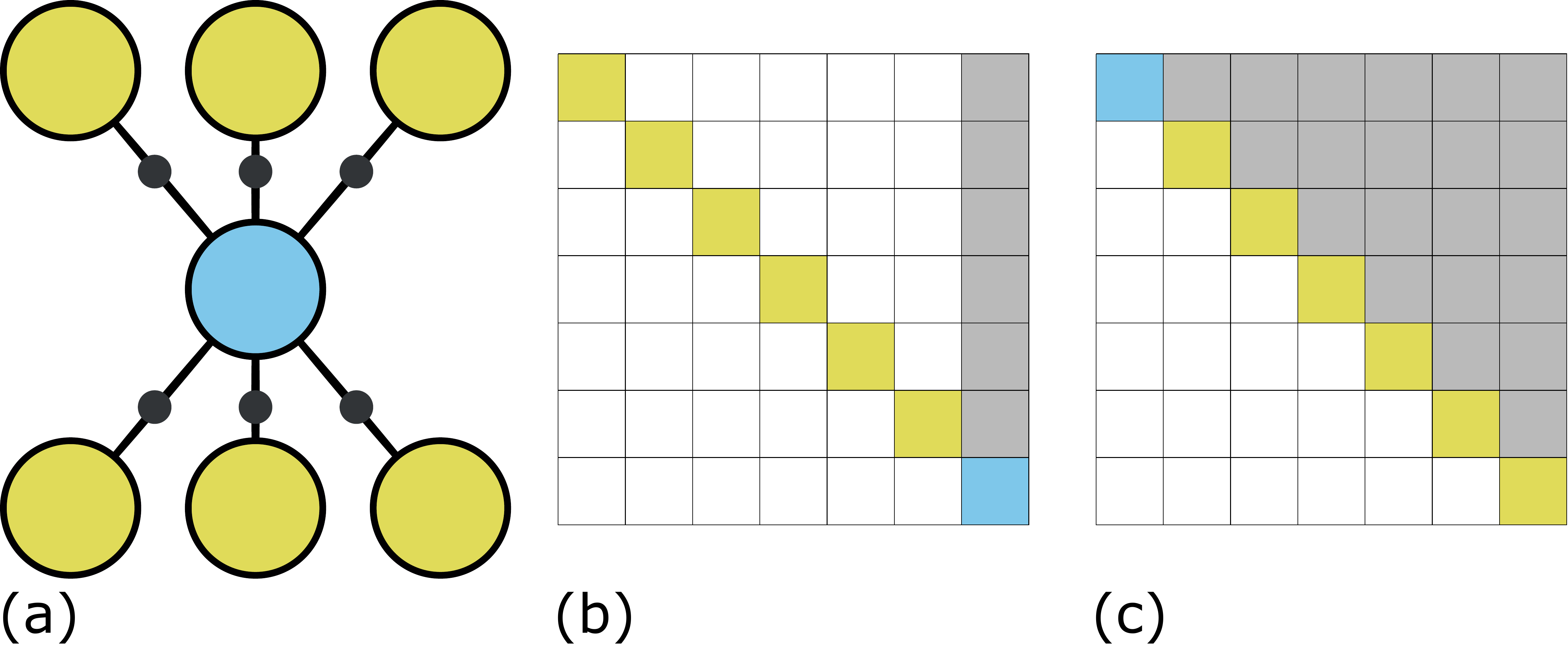}
\caption{(a) A factor graph representing a small belief; the graph contains a "highly connected" variable (in blue). We can examine the sparsity pattern of the belief's square root matrix: (b) shows the sparsity pattern, if the "highly connected" variable is placed last in the variable order; (c) shows the pattern if the variable is placed first. As a rule of thumb, to reduce fill-in, "highly connected" variables should be placed forwards.
}
\label{fig:density}
\end{figure}

First, taking this aspect into account, we shall reconsider the rules for class separation. As a rule of thumb, to reduce fill-in, a "highly connected" variable should be placed forwards, after the variables it is connected to; placing it behind these variables (i.e., eliminating it first, which is equivalent to its marginalization from the belief) would lead to a dense block, involving all the connected variables. This property, which is demonstrated in Fig.~\ref{fig:density}, is the key to "minimum degree" fill-reducing ordering strategies. 
Thus, considering a certain update, having a "highly connected" uninvolved variable pushed back, outside of the affected block, can make that affected block denser, and actually decrease the re-factorization efficiency. \linebreak To avoid such scenarios, we want to identify these "highly connected" variables, and push them further forwards in the variable order, by "bumping" them to higher classes. We thus consider the following adjustment to the original involvement-level-based class separation rule: if a variable has more factors connecting it to (variables in) higher classes, than factors connecting it to (variables in) its own class or lower ones, then forcefully bump it to a higher class. We then mark this fill-optimized variable reclassification as $*\text{class}$.

\newcommand{\AlgEssential}[1]{{\color{teal} #1}}
\newcommand{\AlgOptional}[1]{{\color{teal} #1}}
\newcommand{\AlgFlag}[1]{{\color{RedViolet} #1}}

\begin{algorithm}
\caption{$\pivot$}
\label{alg:pivot} 
\SetKw{return}{return}
\SetKwBlock{inputs}{Inputs:}{}
\SetKwBlock{outputs}{Output:}{}

\small{

\nonl\inputs{
An initial belief $b$ (with state vector $\bm X$), and a set $\mathcal{H}$ of future hypotheses, as described in Section~\ref{sec:pivot-push} \\
$c$ (chosen number of classes), \\
\AlgFlag{FILL\_AWARE} (boolean flag),\\
\AlgFlag{FORCE\_INCREMENTAL} (boolean flag)\\
}
\nonl\outputs{
A permuted state vector $\tilde{\bm X}$ \\
}

\AlgEssential{\tcp{calc variable involvement levels}}
class,$\mathtt{level} \leftarrow zeros(1,n)$ \tcp{init zero vectors} 
\ForAll{$x\in\mathcal{X}$}{
\ForAll{$H\in\mathcal{H}$}{
\If{$x\in\inv_\mathcal{X}(H)$}{
$\mathtt{level}(x) ++$\\
}}}

\AlgEssential{\tcp{divide involved vars into $c$ classes}}
\If{$c = 1$}{
\ForAll{$x\in\mathcal{X}$}{
class$(x) \leftarrow \max\{1, \mathtt{level}(x)\}$\\
}
}
\ElseIf{$c = \max$}{
class$(x) = \mathtt{level}(x)$\\
}
\Else{
$M \leftarrow \max_{x\in\mathcal{X}}\{\mathtt{level}(x)\}$ \\
\ForAll{$x\in\mathcal{X}$}{
class$(x) \leftarrow \min \left\{i\in[0,\dots,c] \mid \mathtt{level}(x) \leq i\cdot\frac{M}{c} \right\}$ \\
}
}
\If{\AlgFlag{FILL\_AWARE = false}}{
\AlgEssential{\tcp{calc $\pivot$, sort vars by class}}
$ \tilde{\bm X} \leftarrow \mathtt{sort}(\bm X,\text{class})$ \\
}
\Else{
\AlgOptional{\tcp{calc fill-aware $\pivot^*$}}

$*$class $\leftarrow$ class\\
\If{\AlgFlag{FORCE\_INCREMENTAL = true}}{
\AlgOptional{\tcp{from first involved var}}
$j_\text{first} \leftarrow \min \left\{j\in[1,\dots,n] \mid \mathtt{level}(x_j) > 0 \right\}$  \\
$*$class$(x_1,\dots,x_{j_\text{first}-1}) \leftarrow$ -1}

\Else{
\AlgOptional{\tcp{from first state var}}
$j_\text{first} \leftarrow 1$
}

\AlgOptional{\tcp{fill-aware var re-classification}}
$\bm \Lambda \leftarrow \bm J^T \bm J$ \\
\ForAll{$j \in [j_\text{first},\dots,n]$}{
\tcp{count how many factors connect $x_j$ to each class}
connections $\leftarrow zeros(1,c)$ \\
\ForAll{$i \in [1,\dots,j-1,j+1,\dots,n]$}{
\If{$\bm \Lambda(i,j) \neq 0$}{
\tcp{count connection between $x_j$ to the class of $x_i$}
connections$(*$class$(x_i)) ++$
}
}
current\_class $\leftarrow$ class$(x_j)$ \\
\While{$\sum \text{connections}(1:\text{current\_class}) < 
		\sum \text{connections}(\text{current\_class}+1:c)$}{
\tcp{bump $x_j$ to a higher class}
current\_class $++$ \\
}
$*$class$(x_j) \leftarrow$ current\_class\\
}

\AlgOptional{\tcp{optimize order in each class}}
new\_order $\leftarrow \mathtt{CCOLAMD}(\bm J, *\text{class})$\\

\If{\AlgFlag{FORCE\_INCREMENTAL = true}}{
new\_order$(1:j_\text{first}-1) \leftarrow [1:j_\text{first}-1$] \\
}
$ \tilde{\bm X} \leftarrow \bm X(\text{new\_order})$
}

}\end{algorithm}

\pagebreak

As another treatment for the fill-in concern, we may combine $\pivot_c$ with a fill-reducing order, such as $\mathtt{CCOLAMD}$. This order is the constrained version of the famous $\mathtt{COLAMD}$ order, which also allows to enforce the relative order among groups of variables, by specifying a certain value $\in\mathbb{N}$ ("constraint") for every variable. Then, for example, all variables with value~$1$ will appear before variables with value~$2$, which will appear before variables with value~$3$, and so on. The internal ordering of variables with the same constraint will match the fill-reducing $\mathtt{COLAMD}$. In our case, intuitively, the variable constraints should match the connectivity-aware variable involvement classes, which we marked $*\text{class}$.

Overall, combining these two ideas yields the fill-aware $\pivot_c$ order, which we mark $\pivot^*_c$:
\begin{multline}
\pivot^*_c(\bm X) \doteq \mathtt{CCOLAMD}(\bm J, \mathtt{constraint}), \\ \text{where }\,
\mathtt{constraint}(x) \doteq i, \,\text{such that }\, x\in*\text{class}^i_{\mathcal{X}}.
\end{multline}
The explicit steps for calculating $\pivot^*_c$ are provided in Algorithm~\ref{alg:pivot}, when the "fill-aware" flag is set to "true".


\subsection{Forced-Incremental PIVOT \label{sec:pivot-inc}}
Maintaining the relative order in the "uninvolved" class, as occurs in $\pivot_c$, means that if the current variable order (i.e., $\bm X$) begins with uninvolved variables, they would keep their position, and not be reordered, i.e.,
\begin{equation}
\big[\pivot_c(\bm X) \big](1,\dots,j-1) \equiv \bm X(1,\dots,j-1),
\end{equation}
where
\begin{equation}
j \doteq \min_{x\in\mathcal{X}}\{\mathtt{index}(x,\bm X) \mid x\in\inv_{\mathcal{X}}(\mathcal{H}) \}
\end{equation}
marks the index of the first involved variable, in any of the updates. In accordance to Section~\ref{sec:applying-order}, this property means that the $\pivot_c$ order can often be \emph{applied} incrementally, with only partial modification of the belief.

\begin{figure}[t]
\includegraphics[width=\textwidth]{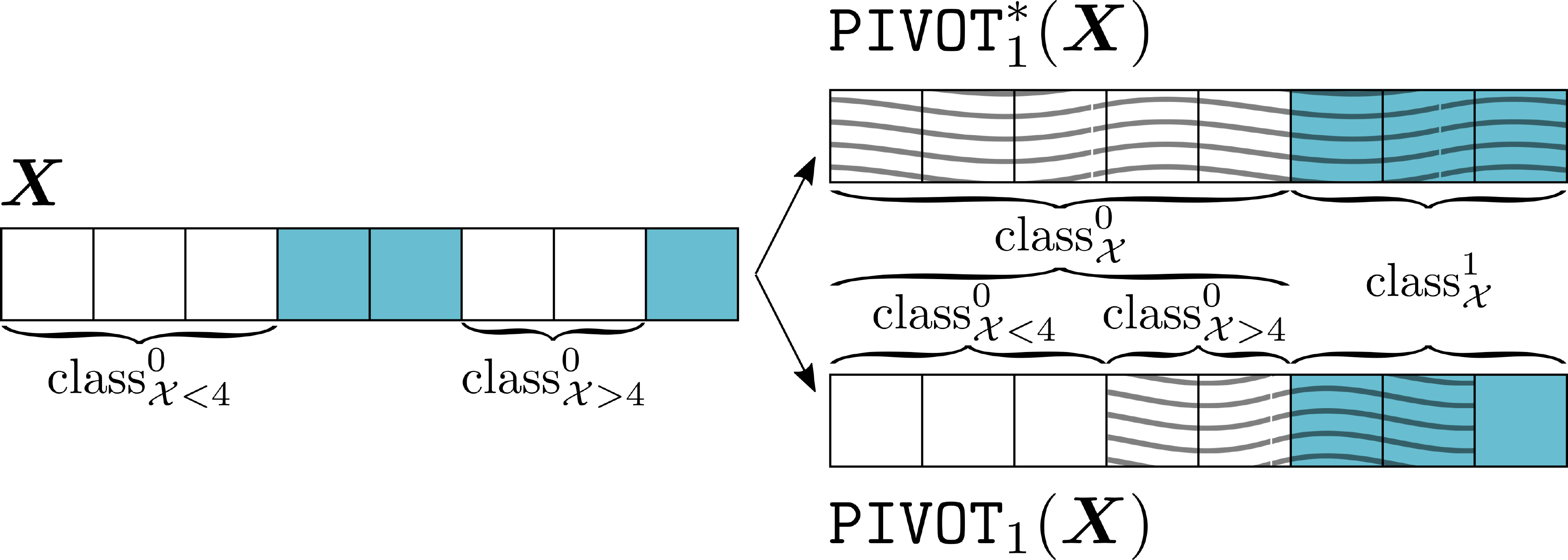}
\caption{Incremental aspects of $\pivot$ vs. $\pivot^*$. On the left -- the initial state vector, where white cells represent uninvolved variables, and blue cells represent involved variables. On the right -- two modified orders, returned from $\pivot_1$, and its fill-aware counterpart $\pivot^*_1$. $\pivot_1$ does not change the order of uninvolved variables that appear at the beginning of the state (before the first involved variable at index~$4$); it can thus be applied incrementally. On the other hand, $\pivot^*_1$ may cause reordering of all variables (black overlay), and must be adjusted to allow incremental application.}
\label{fig:incremental-reordering}
\end{figure}

This property (which is demonstrated in Fig.~\ref{fig:incremental-reordering}) is especially important for large systems, where performing "full" reordering may not be feasible in real time. However, it is not necessarily kept in the fill-aware variation of $\pivot$, where variables in each class might be "shuffled" to a fill-reducing order. If we want to \textbf{make sure the returned order can still be applied incrementally}, we should consider a slight adjustment to it. To this end, we shall first divide $\text{class}^0_{\mathcal{X}}$ (the uninvolved variables) into two sub-classes:
\begin{equation}
\text{class}^0_{\mathcal{X}} =
\text{class}^0_{\mathcal{X}<j} \cup \text{class}^0_{\mathcal{X}>j}
\end{equation}
where
\begin{equation}
\text{class}^0_{\mathcal{X}<j} \doteq \{ x\in \text{class}^0_{\mathcal{X}} \mid \mathtt{index}(x,X) < j \}
\end{equation}
contains all the uninvolved variables that already appear at the beginning of the state, before the first involved; and 
\begin{equation}
\text{class}^0_{\mathcal{X}>j} \doteq \{ x\in \text{class}^0_{\mathcal{X}} \mid \mathtt{index}(x,X) > j \}
\end{equation}
contains all the uninvolved variables that appear after the first involved, and should be pushed back. This division is also demonstrated in Fig.~\ref{fig:incremental-reordering}.

Then, to avoid affecting the order of the first $j$ variables, our fill-aware optimizations should be adjusted as follows: first, the reclassification of "highly connected variables" should ignore the variables in $\text{class}^0_{\mathcal{X}<j}$; second, when applying $\mathtt{CCOLAMD}$, the constraint value used for the variables in $\text{class}^0_{\mathcal{X}<j}$ should be set to $-1$, to force them backwards; third, after the application of $\mathtt{CCOLAMD}$, we should "cancel" the internal ordering of those variables, by forcing:
\begin{equation}
\big[\pivot^*_c(\bm X) \big](1,\dots,j-1) = \bm X(1,\dots,j-1) 
\end{equation}
These steps are explicitly formulated as part of Algorithm~\ref{alg:pivot}, when the "force-incremental" flag is set to "true".

Note that, although such incremental application of $\pivot^*_c$ is more efficient, it can lead to slightly inferior fill reduction in comparison to the standard counterpart. Still, the sub-optimal fill-in in the "uninvolved" rows does not affect the efficiency of the updates, by definition.

%% file: pivot-for-inference.tex
\subsection{\changedRemoved{The Influence of PIVOT on State Inference and Re-Planning \label{sec:influence-on-inf}}}
Consider that in the planning session at time-step $k$, we applied $\pivot$ on the initial belief. After policy selection, we have the option to continue to the following inference and planning sessions with the belief in the original order (assuming it was cached), or continue with the belief in the $\pivot$ order. Let us examine the benefits (and possible shortcomings) of continuing with the modified order.

First, consider another planning session is due in the future.
Since $\pivot$ can be applied incrementally, if the $\pivot$ order is kept after the current planning session, then in the next planning session, a "new" $\pivot$ order can simply be incrementally updated, according to the change in variable classification (and addition of new variables). Assuming the classification of uninvolved variables usually does not change much between consecutive planning sessions, maintaining the $\pivot$ order would gradually increase the size of the "fixed" variable class $\text{class}^0_{\mathcal{X}<j}$, and reduce the cost of re-applying $\pivot$ in the future.

Second, we can examine the influence on future state inference updates, due when executing the selected policy. As explained, incorporating updates to variables from the past (i.e., loop closures) is expensive, when these affect numerous variables. When applying $\pivot$ during a planning session, we push forwards predicted loop closing variables, which we believe to be re-observed when executing the policy. We explained how doing so can reduce the computational cost of performing the predicted updates during planning. Now, if we are to execute the selected policy, it is likely that we indeed face the same loop closures we predicted, and need to perform similar belief updates. Hence, in this case, keeping the predicted loop closing variables forwards (i.e., maintaining the $\pivot$ order after planning) can reduce the computational cost of the following state inference updates, just like it did during the (predicted) planning updates. As mentioned, when considering "pure" state inference, and ignoring the planning problem, there is no point in actively pushing forwards loop closing variables, as the cost of such reordering is equivalent to actually performing the loop closure. However, since we conduct the reordering procedure anyway, as part of the planning, we gain this indirect benefit to the state inference process. This conclusion is derived from a wider view of the system, which considers both the state inference and planning processes.


Finally, we should also consider that $\pivot$ is intended to increase planning efficiency, in the ways indicated, and is not inherently fill-reducing. Typically, maintaining the $\pivot$ variable order after planning, would result in a sub-optimal fill-in in the belief during state inference, and increase the memory footprint. Also, although the cost of inference updates should actually improve, as we explained, the added fill-in in the square root matrix might increase the cost of back-substitution, which is performed when updating the state MAP estimate. The frequency of back-substitution in the system is a matter of design. For example, one may compute it after every state inference session, or only before planning. We also do not always need to calculate the entire back-substitution solution; updating the estimate of the last pose, which is usually at the end of the state, can be achieved almost immediately. Thus, overall, even if we expect some sacrifice in fill-in and back-substitution cost, we expect the gains in performance to prevail. 

%% file: results.tex
\section{\changed{Experimental Evaluation \label{sec:results}}}
\graphicspath{ {./images/scenario/} }
\changed{Next, we wish demonstrate the advantages of $\pivot$, and present an in-depth analysis of the approach, in a realistic active SLAM simulation. In this simulation, a ground robot is tasked to autonomously navigate through a sequence of eight pre-defined goal points in an unknown indoor environment. The robot knows the location of the goals, but must perform SLAM, in order to accurately maintain its own location in the environment (i.e., localize). To maintain safety, and make sure the goal points are indeed reached, the robot should actively reason about its future belief, in order to choose the most informative path to each goal, which minimize the uncertainty over its trajectory. For example, when choosing a path which passes through a previously explored area, certain loop closures are expected to be added to the belief, reducing its uncertainty; hence, different paths might lead to different loop closures, and convey different uncertainty levels.}

\subsection{Simulation Overview}
\begin{figure}[t]
	\includegraphics[width=0.49\textwidth]{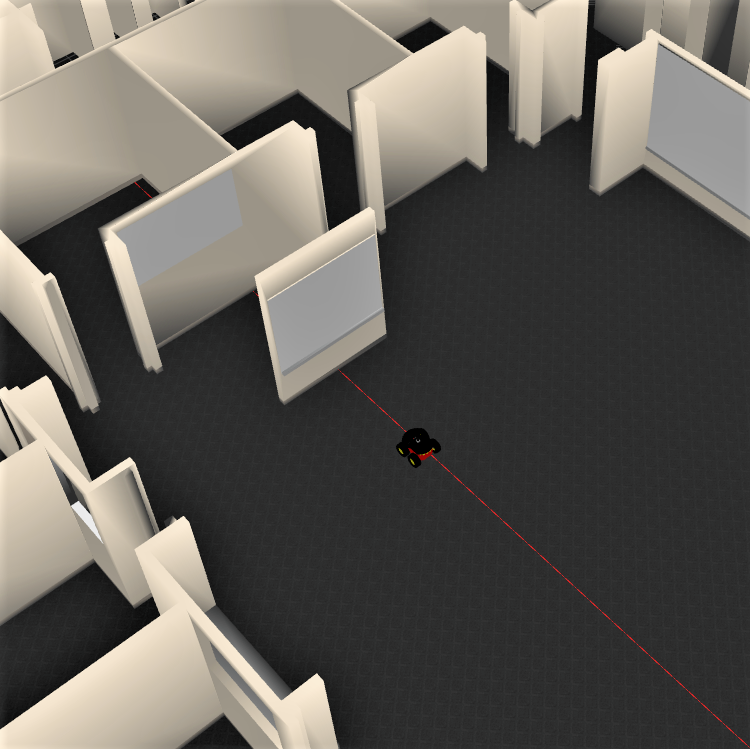}
	\includegraphics[width=0.49\textwidth]{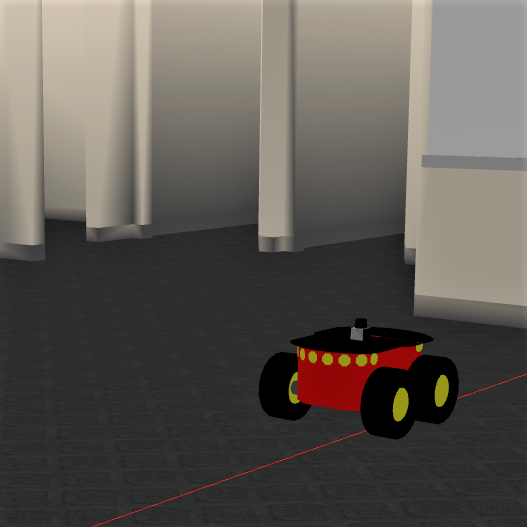}
	\caption{A Pioneer 3-AT robot in the simulated indoor environment. The robot is equipped with a lidar sensor, Hokuyo UST-10LX, as visible on top of it.}
	\label{fig:scenario-robot}
\end{figure}

We used the Gazebo simulation engine \citep{Koenig04iros} to simulate the environment and the robot (Pioneer 3-AT), which is equipped with a lidar sensor (Hokuyo UST-10LX), providing point-cloud observations; these components can be seen in Fig.~\ref{fig:scenario-robot}. A grid-based map of the environment is incrementally approximated by the robot from the scans during the navigation. Robot Operating System (ROS) is used to run and coordinate the system components -- state inference, decision making, sensing, etc. The robot's state $\bm X_k \doteq [x^T_0,\dots,x^T_k]^T$ consists of discretized poses kept along its entire executed trajectory. Its belief over this trajectory is represented as a factor graph, and maintained using the GTSAM C++ library \citep{Dellaert12tr}. Each pose $x_k$ consists of three variables, representing a 2D position and orientation (in a global coordinate frame). The robot's initial pose is provided to it as a prior factor over $x_0$ (with white Gaussian noise). After passing a set distance, a new pose is added to the end of the state, and a lidar scan (in a range of approximately 30 meters) is taken. A motion constraint (pose transition factor), with white Gaussian noise, matching the real hardware's specs, is added between the new pose and the previous one. The observed point-cloud is then matched to scans taken from key past poses using ICP matching \citep{Besl92pami}. If a match is found, a Gaussian loop closing constraint (factor) is added between these poses. After adding the new motion and observation constraints, the agent incrementally updates its belief and state estimate; we refer to this belief update process as a "state inference session". To follow a path, the robot uses a "pure pursuit" controller \cite{Coulter92report}, which yields an appropriate sequence of control actions, based on the robot's estimated position. To initiate the simulation, and build a rudimentary map, the robot is set to follow a short predefined path, given as a sequence of way-points.

\subsection{Planning}
After completing the initial predefined path, the robot performs a "planning session", based on its latest belief $b$, towards the first goal. It begins by generating a set $\Pi$ of twenty collision-free candidate paths (sequences of way-points) from its estimated pose to the goal: it first execute the Probabilistic RoadMap (PRM) algorithm \citep{Kavraki96tra} to sample the map, and then executes the K-diverse-paths algorithm \citep{Voss15icra} on the sampled PRM graph, to return a set of topologically diverse paths.

To evaluate and compare the "safety" of taking each path, we use the previously-defined belief-based information-theoretic value function $\tilde{V}$ (from (\ref{eq:objective})), which measures the expected information gain. For tractability, we evaluate each candidate by predicting only the ML future hypothesis, as explained in Section~\ref{sec:prelim-def}. Hence, each candidate path $\pi\in\Pi$ is pre-matched with a single factor graph $FG^\delta_\pi$, containing the factors and poses to be added to the prior belief graph $FG$, after following the path. For each candidate, loop closure factors are predicted between future and past poses, if their estimated location is within a close range (i.e., where we expect to add them when following this path); thus, we do not need predict the raw laser scans, but only which poses would be connected with a factor. 

\begin{figure}[t]
	\begin{subfigure}[t]{0.9\textwidth}
        \includegraphics[width=\textwidth]{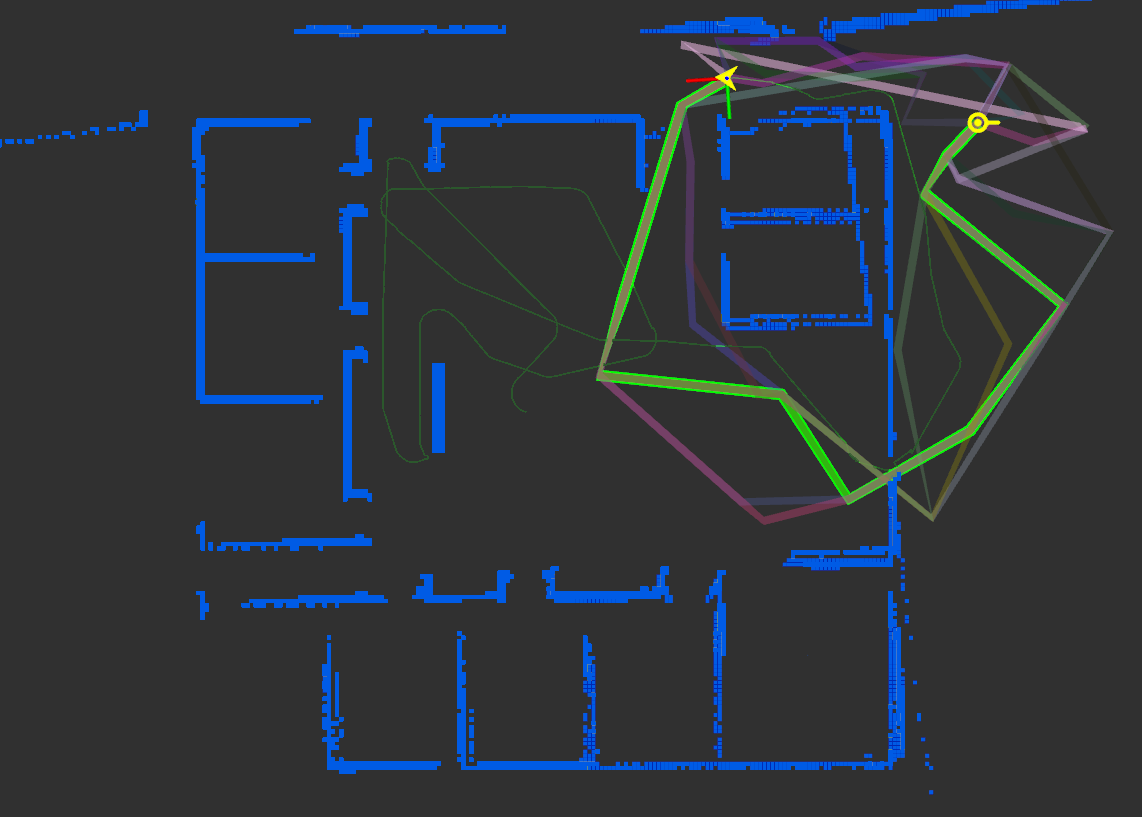}
        \caption{Planning session \#1.}
	\end{subfigure}
	\begin{subfigure}[t]{0.9\textwidth}
        \includegraphics[width=\textwidth]{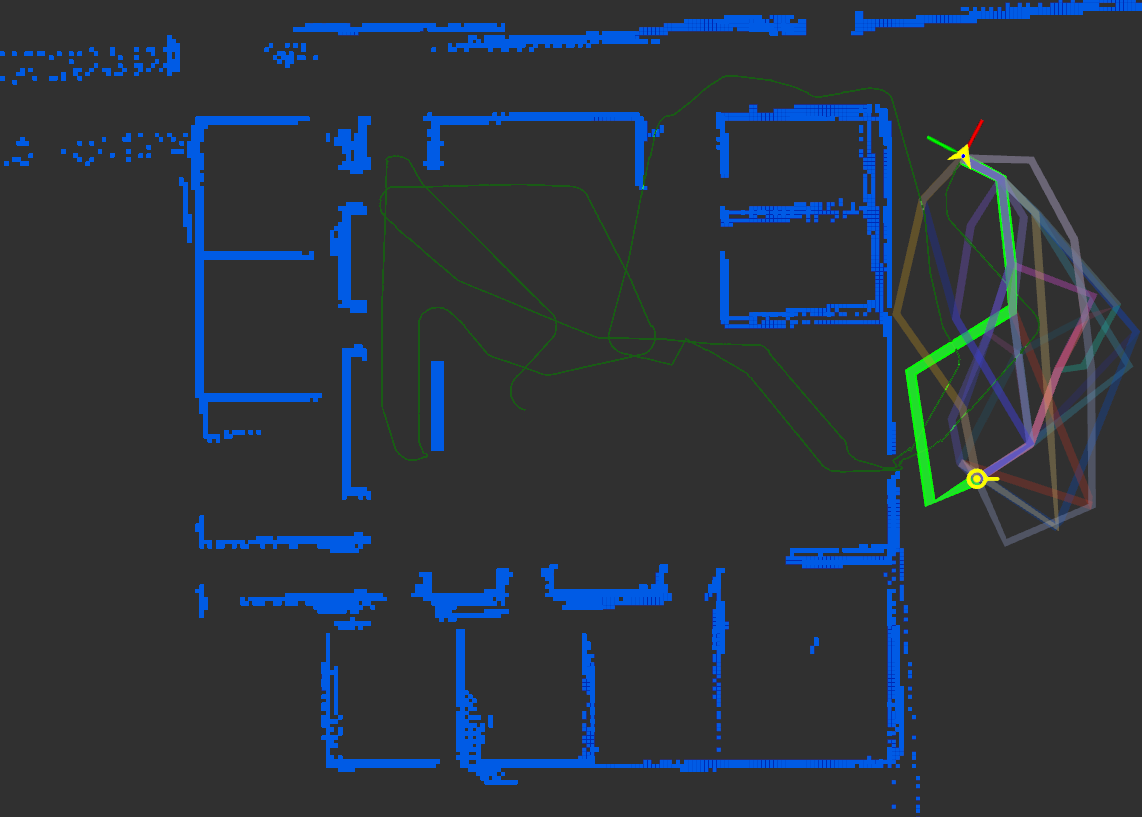}
        \caption{Planning session \#2.}
	\end{subfigure}
    \begin{subfigure}[t]{0.9\textwidth}
        \includegraphics[width=\textwidth]{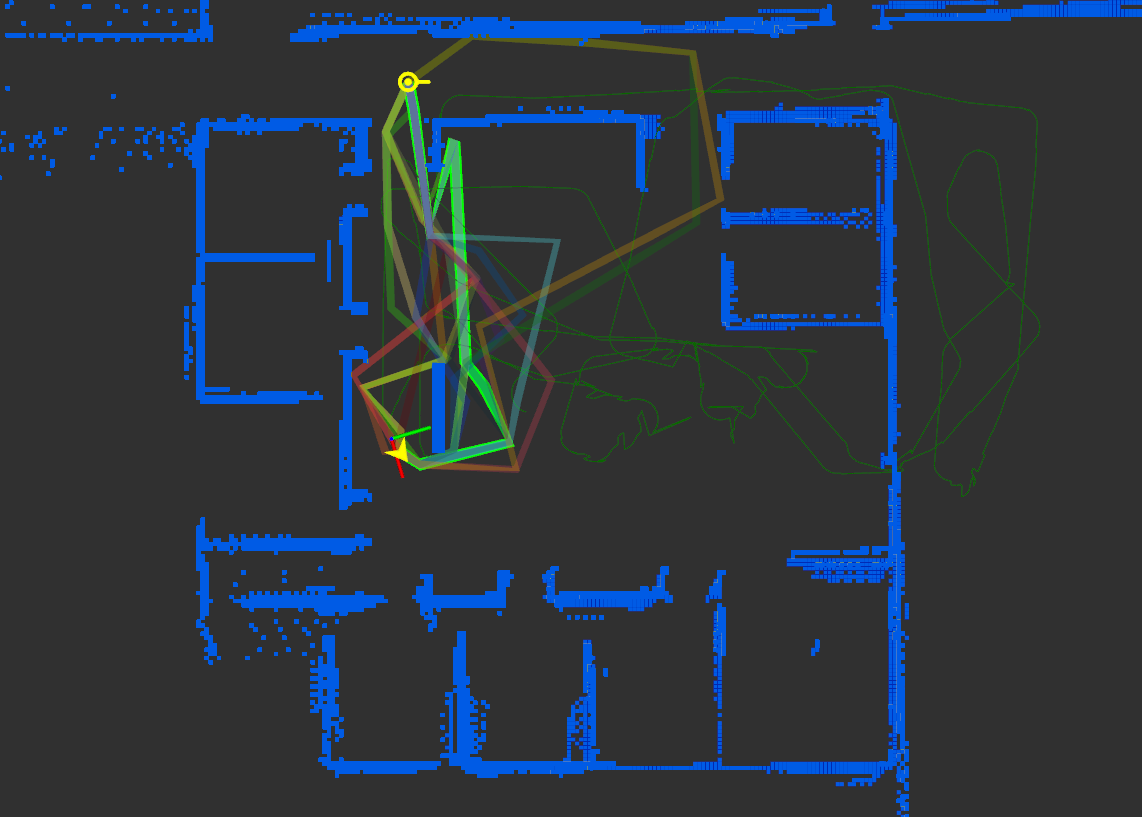}
        \caption{Planning session \#7.}
    \end{subfigure}
\caption{Visualization of the planning sessions, including: the estimated map (blue occupancy grid); the current estimated pose (yellow arrow-head) and goal (yellow circle); the trajectory taken up to that point (thin green line); the candidate trajectories from the current position to the goal (thick lines in various colors); and the selected trajectory (highlighted in bright green). The size of the mapped area "in reality" is (roughly) 20m$\times$30m.}
    \label{fig:planning-sessions}
\end{figure}

After updating the initial belief according to each of the candidate hypotheses, the value of each candidate is computed, and the optimal path is selected (marking the end of this planning session), and then followed. After reaching the goal at the end of the selected path, the robot conducts a new planning session, towards its next goal, until all goals are reached. In total, the robot conducts eight planning sessions, separated by sequences of state inference sessions. A visualization of (some of) the planning sessions is provided in Fig.~\ref{fig:planning-sessions}.

\subsection{\changedRemoved{Comparing the Performance of PIVOT} }
In the considered scenario, the state size grows quickly as the navigation progresses, making planning more computationally challenging. Hence, we wish to examine how utilizing $\pivot$ can help reducing the planning cost. To do so, throughout the simulation, we maintained in parallel multiple versions of the (factorized) belief, using different ordering tactics, and performed the appropriate belief updates on all versions of the belief, during each planning and state inference session. We recall that the different ordering tactics only convey a change of representation, and do not affect the accuracy of the solution.

The baseline ordering tactic we considered matches the one suggested by the state-of-the-art SAM algorithm iSAM2 \citep{Kaess12ijrr}. 
According to this tactic, at each incremental update of the factorization, we shall calculate and apply a constrained fill-reducing order ($\mathtt{CCOLAMD}$) on the affected variables; the constraint forces the involved variables forward, to the end of the state. While the reasons for applying a fill-reducing order as part of this tactic are obvious (reducing fill-in and the update cost), the added constraint actually leads to a sacrifice in fill-in, and does not contribute to the current inference session. However, the constraint is added to contribute to the following inference session: if the variables are to be updated again, then the affected block in the next session would be smaller. This tactic is often worthwhile for incremental inference.

We also utilized several variations of $\pivot$ during planning, which meant performing additional standalone reordering of the belief before calculating the objective values in each planning session. This comes in contrast to the baseline tactic, where (incremental) reordering is only performed during updates, and never as a standalone step. For comparison, we considered $\pivot_1$, $\pivot_5$, and $\pivot_{\max}$, which are the basic $\pivot$ orders, with varying numbers of classes, and their fill-aware counterparts $\pivot^*_1$, $\pivot^*_5$, and $\pivot^*_{\max}$. Although it is not obligatory, for this comparison, we chose to always continue to the following state inference sessions (after the planning is completed) with the modified orders; nonetheless, all updates during the state inference sessions were always conducted according to the baseline tactic.

Overall, we maintained seven versions of the belief, according to seven ordering tactics: the baseline -- in which we performed reordering only during updates, and six other tactics, in which we additionally applied a $\pivot$ variation in standalone reordering steps during planning sessions. For each version, we measured and compared the total state inference and planning times. For each inference session, we measured the time of performing the belief factorization update and the back-substitution (to refine the MAP estimate); for each planning session, we measured the belief update time according to the predicted hypothesis of each candidate, along with the time of applying the $\pivot$ order, when it was used. As explained in Section~\ref{sec:applying-order}, applying the order requires partial re-factorization; to this end, we used the modification algorithm developed in our previous work \citep{Elimelech21ral}.

For a fair comparison, we chose to the detach the timing of the belief update and reordering processes from any framework-specific implementation overhead. As explained in Section~\ref{sec:prelim-square-root-sam}, since the system is Gaussian, the belief update process can be performed by updating the square root information matrix. Thus, at each state inference session, we extracted the new Jacobian rows from our GTSAM factor graph, representing the newly added (linearized) factors. We then performed the incremental update of the belief's square root matrix, in "iSAM2 fashion", using a standard MATLAB implementation \citep[see][]{Elimelech21thesis}. At each planning session, we extracted the Jacobian rows representing the predicted hypotheses, and performed a similar square root matrix update for each one. Of course, the columns of the extracted Jacobian rows were reordered appropriately before each update, to match the variable order of that version of the belief. The sparsity pattern of the square root matrix under different variable orders is presented in Fig.~\ref{fig:scenario-fill-in}.

\begin{figure}[t]
	\begin{subfigure}[t]{\textwidth}
        \includegraphics[width=\textwidth]{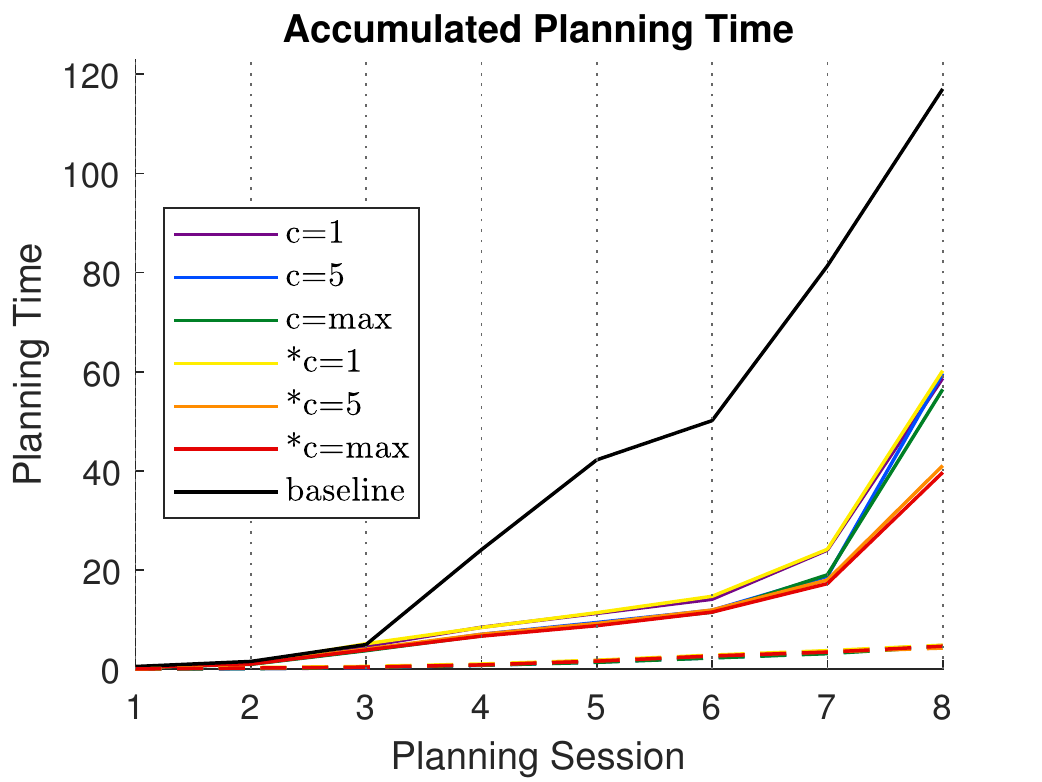}
        \caption{}
        \label{fig:scenario-timing-planning}
	\end{subfigure}
	\begin{subfigure}[t]{\textwidth}
        \includegraphics[width=\textwidth]{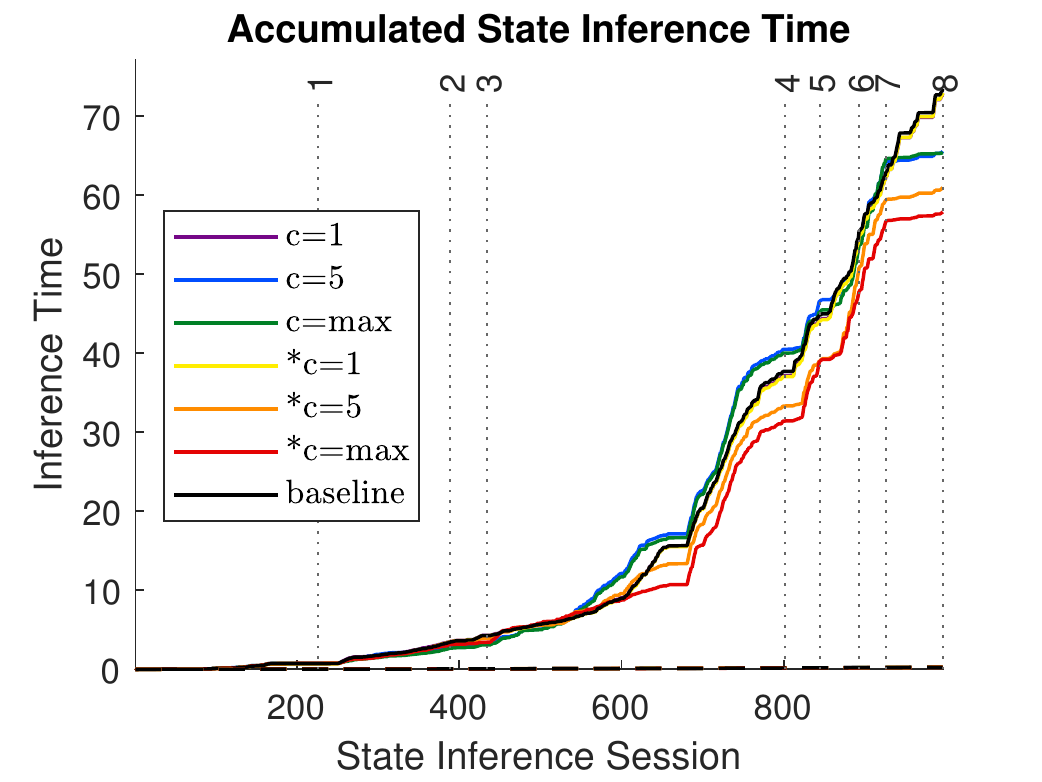}
        \caption{}
        \label{fig:scenario-timing-inf}
	\end{subfigure}
    \begin{subfigure}[t]{\textwidth}
        \includegraphics[width=\textwidth]{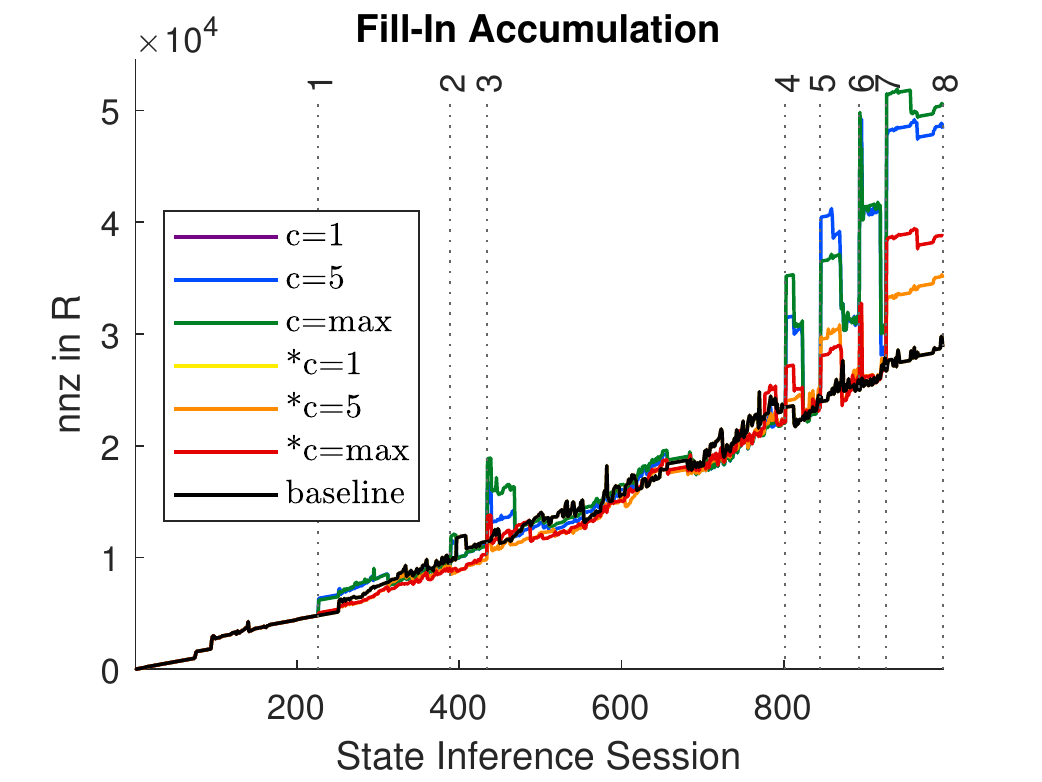}
        \caption{}
        \label{fig:scenario-fill-in-per-session}
    \end{subfigure}
\caption{Comparison of the experiment solution, using different variable ordering tactics: baseline ("iSAM2"), three variants of $\pivot_c$, and three variants of $\pivot^*_c$.\\
(a) Accumulated planning time throughout the eight planning sessions; lower is better. The dashed line represents the accumulated reordering time out of the measured time.\\ 
(b) Accumulated state inference time throughout the ($\sim$)1,000 inference sessions; lower is better. The dashed line represents the accumulated back-substitution time out of the measured time.
(c) Fill-in accumulation (number of non-zeros in $\bm R$).}
    \label{fig:scenario-timing}
\end{figure}

\begin{table*}[ht]
\footnotesize
\begin{tabularx}{\textwidth}{|l|l?X?X|X|X?X|X|X|}
\hline 
\multicolumn{2}{|l?}{Tactic}
& Baseline
& \multicolumn{3}{c?}{$\pivot_c$}  
& \multicolumn{3}{c|}{$\pivot^*_c$} \\
\hline
\multicolumn{2}{|l?}{\# of classes}
& N/A
& $1$ 
& $5$ 
& $\max$ 
& $1$ 
& $5$ 
& $\max$ 
\\\Xhline{5\arrayrulewidth}

\multirow{2}{*}{Planning Session \#1} 
& Reordering time & / & 00.05 & 00.06 & 00.06 & 00.07 & 00.08 & 00.07
\\\cline{2-9}
&  Update time    & 0.56 & 00.39 & 00.30 & 00.30 & 00.26 & 00.29 & \textbf{00.28}
\\\Xhline{3\arrayrulewidth}

\multirow{2}{*}{Planning Session \#2} 
& Reordering time & / & 00.11 & 00.06 & 00.06 & 00.26 & 00.20 & 00.16
\\\cline{2-9}
&  Update time    & 01.02 & 00.64 & 00.64 & 00.56 & 00.57 & 00.52 & \textbf{00.52}
\\\Xhline{3\arrayrulewidth}

\multirow{2}{*}{Planning Session \#3} 
& Reordering time & / & 00.05 & 00.21 & 00.22 & 00.23 & 00.18 & 00.18
\\\cline{2-9}
&  Update time    & 03.40 & 03.36 & \textbf{02.52} & 02.53 & 03.72 & 02.78 & 02.64
\\\Xhline{3\arrayrulewidth}

\multirow{2}{*}{Planning Session \#4} 
& Reordering time & / & 00.63 & 00.45 & 00.45 & 00.48 & 00.42 & 00.45
\\\cline{2-9}
&  Update time    & 19.12 & 03.20 & 02.77 & 02.56 & 02.78 & 02.54 & \textbf{02.33}
\\\Xhline{3\arrayrulewidth}

\multirow{2}{*}{Planning Session \#5} 
& Reordering time & / & 00.90 & 00.66 & 00.59 & 00.76 & 00.79 & 00.75
\\\cline{2-9}
&  Update time    & 18.12 & 01.88 & 01.71 & 01.71 & 02.22 & 01.37 & \textbf{01.35}
\\\Xhline{3\arrayrulewidth}

\multirow{2}{*}{Planning Session \#6} 
& Reordering time & / & 00.84 & 00.85 & 00.88 & 01.02 & 01.02 & 01.03
\\\cline{2-9}
&  Update time    & 07.90 & 02.00 & \textbf{01.61} & 01.54 & 02.27 & 01.69 & 01.70
\\\Xhline{3\arrayrulewidth}

\multirow{2}{*}{Planning Session \#7} 
& Reordering time & / & 01.01 & 00.92 & 00.86 & 00.89 & 00.79 & 00.78
\\\cline{2-9}
&  Update time    & 31.20 & 08.89 & 05.92 & 06.67 & 08.58 & 05.26 & \textbf{04.95}
\\\Xhline{3\arrayrulewidth}

\multirow{2}{*}{Planning Session \#8} 
& Reordering time & / & 01.15 & 01.40 & 01.42 & 01.09 & 00.75 & 01.20
\\\cline{2-9}
&  Update time    & 35.70 & 33.49 & 39.52 & 35.98 & 34.91 & 22.30 & \textbf{21.23}
\\\Xhline{5\arrayrulewidth}

\multirow{2}{*}{Total planning time} 
&  Sec.    & 117.05 & 58.69 & 59.68 & 56.45 & 60.21 & 41.07 & \textbf{39.70}
\\\cline{2-9}
& Relative & 100\% & 50.14\% & 50.98\% & 48.22\% & 51.43\% & 35.08\% & \textbf{33.91\%}
\\\Xhline{5\arrayrulewidth}

\multirow{2}{*}{Total state inference time} 
&  Update time    & 73.19 & 72.60 & 65.27 & 65.08 & 72.69 & 60.71 & \textbf{57.62}
\\\cline{2-9}
& Back-sub. time  & \textbf{00.24} & 00.25 & 00.28 & 00.31 & 00.24 & 00.25 & 00.26
\\\Xhline{5\arrayrulewidth}

\multirow{2}{*}{Total solution time} 
&  Sec.    & 190.49 & 131.55 & 125.24 & 121.82 & 133.16 & 102.05 & \textbf{97.57}
\\\cline{2-9}
& Relative & 100\% & 69.00\% & 65.74\% & 63.95\% & 69.90\% & 53.57\% & \textbf{51.22\%}
\\\Xhline{5\arrayrulewidth}

\multirow{2}{*}{Fill-in (nnz in the $\bm R$ matrix)} 
&  before final reordering    & \textbf{28842} & \textbf{28842} & 48372 & 50472 & \textbf{28842} & 35029 & 38749
\\\cline{2-9}
& after final reordering  & / & 31571 & 62091 & 69376 & \textbf{27662} & 32737 & 40882
\\\hline
\end{tabularx}
\caption{Summary of the accumulated inference and planning times, and final fill-in, as presented in Fig.~\ref{fig:scenario-timing}. Best values in \textbf{bold}.}
\label{tbl:scenario-timing}
\end{table*}

\subsection{Results and Discussion \label{sec:discussion}}
Next, we present and analyze the results of using each ordering tactic to solve the described scenario. Fig.~\ref{fig:scenario-timing-planning} presents the accumulated planning time (update and reordering), using each ordering tactic, throughout the eight planning sessions. Fig.~\ref{fig:scenario-timing-inf} similarly presents the accumulated state inference time (update and back-substitution), using each ordering tactic. Fig.~\ref{fig:scenario-fill-in-per-session} presents the number of non-zero entries in the square root matrix of the current belief, which represents the fill-in of the factorization, in each state-inference session, using each ordering tactic. All the numeric values from these figures are summarized in Table~\ref{tbl:scenario-timing}.

Several conclusions can be made:
first, it is clear to see that using any variation of $\pivot$ managed to substantially reduce the planning time in comparison to baseline tactic ($50-66\%$ reduction), as intended. We can also see that the reordering (applying the $\pivot$ order) time in each planning session becomes minor, in comparison to the update time, as the state grows. Further, all variations of $\pivot$ managed to reduce the state inference time -- verifying our claims in Section~\ref{sec:influence-on-inf} -- and, by such, the total solution time of the scenario ($30-50\%$ reduction). To explain the contribution of $\pivot$ to the inference process, in comparison to the baseline tactic, we may recall that with that tactic we pushed forwards the loop closing variables only \emph{after} they were met; \linebreak when applying $\pivot$, we pushed forwards (predicted) loop closing variables \emph{before} meeting them, in the preceding planning sessions. This resulted in a more efficient loop closing during state inference compared to the baseline.

More specifically, we can see that each variation of $\pivot$ led to different results. Whether examining the per-session or accumulated results, the fill-aware variations of $\pivot$ resulted in lower planning time, inference time, and \mbox{fill-in} (number of non-zeros in $\bm R$), in comparison to their "standard" counterparts; the belief reordering time is also lower in the fill-aware variations. Also, while, generally, using $\pivot/\pivot^*$ led to increase in fill-in, in comparison to the baseline, the fill-aware variants resulted in much lower fill-in, in comparison to their standard counterparts, as expected. These results encourages us to always utilize the $\pivot^*$ over $\pivot$. In most cases, increasing the number of classes (as determined by the parameter $c$), led to lower planning and inference times; yet, with both $\pivot$ and $\pivot^*$, this also resulted in increased fill-in. When using only a single $\pivot$ class (i.e., $c=1$), we can see that the fill-in was hardly impacted. In fact, by applying $\pivot^*_1$ in the final planning session, we even managed to reduce the fill-in below the baseline level. Interestingly, we can see that in the last planning session, the added fill-in induced by $\pivot_5$ and $\pivot_{\max}$ was so significant, that it actually made the planning cost higher, in comparison to the baseline; this might occur when the number of non-zeros in the affected block (of a candidate update) grows more significantly than the decrease in block size. Fortunately, this is not an issue when considering the fill-aware $\pivot^*$ tactics. In fact, it is evident that $\pivot^*_{\max}$ achieved the best performance in both planning and inference.

\begin{figure*}[t]
\begin{subfigure}[t]{0.33\textwidth}\hfill
        \includegraphics[width=0.75\textwidth]{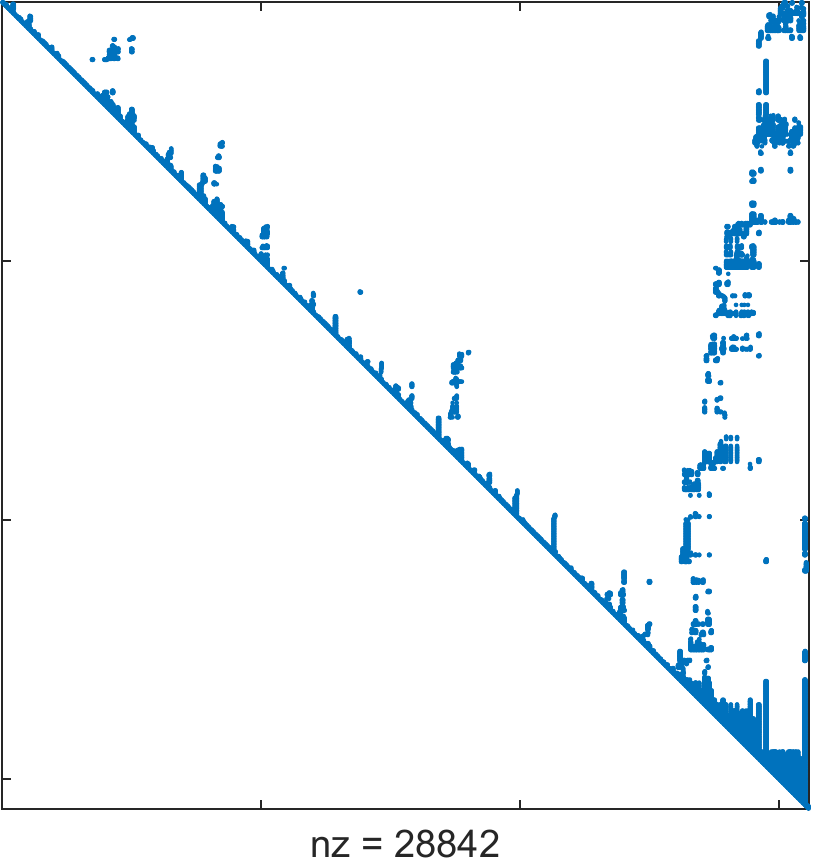}\hspace{20pt}
        \caption{Baseline $\bm R$.}
\end{subfigure}
\begin{subfigure}[t]{0.33\textwidth}\center
        \includegraphics[width=0.75\textwidth]{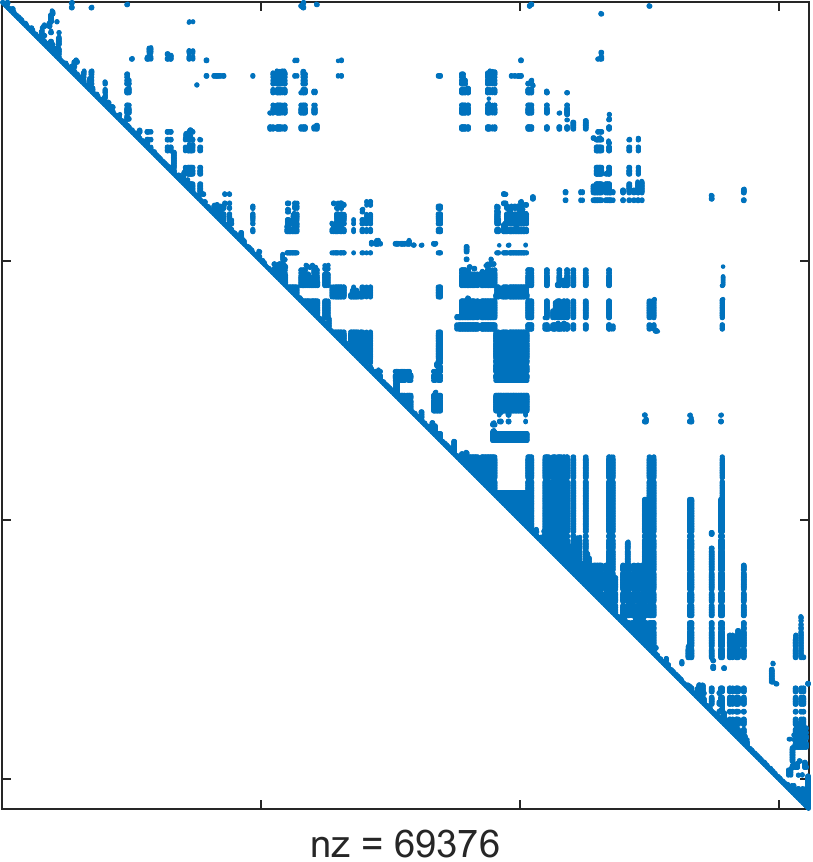}
        \caption{$\bm R$ after the applying $\pivot_{\max}$.}
\end{subfigure}
\begin{subfigure}[t]{0.33\textwidth}
        \hspace{20pt}\includegraphics[width=0.75\textwidth]{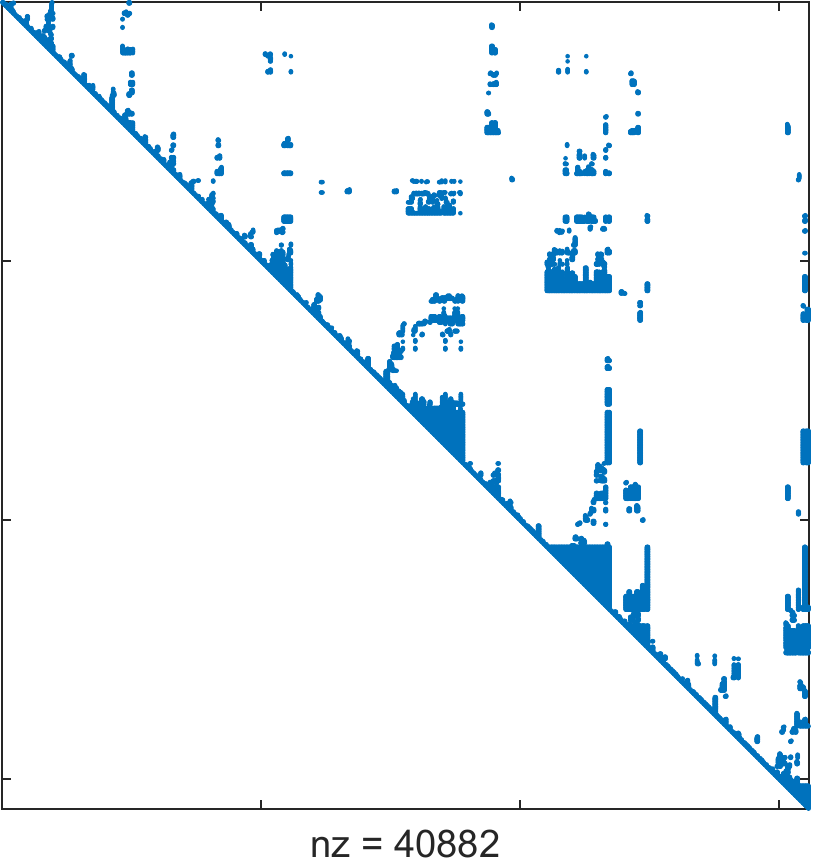}
        \caption{$\bm R$ after the applying $\pivot^*_{\max}$.}
\end{subfigure}
\caption{A comparison of the sparsity pattern of three variants of the agent's square root matrix $\bm R$, at the final planning session.}
    \label{fig:scenario-fill-in}
\end{figure*}

We may also notice that the added fill-in led to a slight increase in the back-substitution time during the inference sessions; yet, the cost of this step is insignificant in relation to the update cost. As mentioned, Fig.~\ref{fig:scenario-fill-in} shows the difference in fill-in in three variations of the square root matrix, at the final planning session: the baseline, after the applying $\pivot_{\max}$, and  $\pivot^*_{\max}$. 

We can see that between planning sessions $3$ and $4$, which were seprated by numerous state inference sessions, the fill-in induced by the $\pivot$ variants decreased back to the baseline level. It appears that the incremental reordering that takes place in those inference sessions has gradually "canceled out" the $\pivot$ constraints, and realigned the order back to the baseline. In contrast, in the latter planning sessions, which occurred more frequently in between the inference sessions, we can see that the fill-in accumulated. Overall, we may conclude that frequent re-planning, and/or a higher number of classes, can amplify both $\pivot$'s disadvantages (relative added fill-in), and benefits (relative reduction in computation time).

As mentioned, we chose to maintain the $\pivot$ order(s) after every planning session. Although we saw this can indeed help reducing the state inference cost, if we wanted, we could also revert back to the original variable order after each planning session. That way, the state inference process (and the fill-in in $\bm R$) would not influenced by $\pivot$. Potentially, we can also analyze the fill-in in the affected block of the selected action, to verify that the state inference is indeed expected to improve by keeping the $\pivot$ order, before making such decision.

Surely, the improvement in planning time tightly depends on the number of candidates, the belief's topology, and the size of the state; the improvement in total solution time is also a parameter of the planning to inference session ratio. Nonetheless, these results encourages us to utilize $\pivot$ in the solution of high-dimensional planning problems.


%% file: conclusion.tex
\section{\changedRemoved{Conclusion \label{sec:conclusion}}}
In this work, we presented a novel idea to improve the efficiency of planning in the belief space, with no sacrifice in accuracy. First, we recognized that the order of state variables of a belief, as expressed in its "square root matrix", determines which subset of them would be affected by future belief updates. We also recognized that, during planning, to evaluate the different candidates, we should perform multiple updates on the same belief, "in parallel". Thus, when facing such a scenario, we suggested to perform a precursory optimization to the variable order, in order to reduce number of affected variables, and the overall planning cost. We referred to this reordering paradigm as PIVOT: Predictive Incremental Variable Ordering Tactic, and presented fill-aware and incremental variations of it. Essentially, PIVOT helps reducing the update cost by bringing forwards "loop closing variables", and sparing the update of "uninvolved variables", as identified by the candidate updates. 

We also saw that as a "by-product" of applying the tactic, we are able to improve the efficiency of the state inference sessions, which are due when executing the selected policy: if we maintain the PIVOT order after the planning session, and our prediction is in-line with the ground truth during execution, then we would similarly reduce the cost of loop closures, when they actually occur.

To demonstrate the effectiveness and benefits of the approach, we applied PIVOT in a realistic active SLAM simulation. There, we solved a sequence of planning and state inference sessions, using various variable ordering tactics. We proved that, using PIVOT, we were able to significantly reduce the computation time of both the planning and inference, with only a slight increase in fill-in (the memory footprint), and with no sacrifice in accuracy.

Since PIVOT does not make any assumptions on the belief structure, as future work, it would be interesting to also verify its benefits when applied to other belief topologies, such as in full-SLAM (vs. pose-SLAM), or when considering non-binary factors. We may also be able to further optimize the order, by utilizing more advanced fill-reducing concepts, such as graph dissection \citep{Krauthausen06rss}, or the Bayes tree \citep{Kaess10tr}. Finally, we would also like to examine the applicability of PIVOT to other contexts, such as multi-hypothesis or multi-modal beliefs \citep{Pathak18ijrr}, where we might also face branching belief updates.

%% file: appendix-obs.tex
\section{\changed{Generalization to Multiple Hypotheses\label{sec:appendix-general-obs}}}
\subsection{Predicting Observations in High-Dimensional States\label{sec:predicting-obs}}
As explained, for each candidate policy, we seek to calculate the expected return (i.e., value function $V$), considering the possible predictable hypotheses for the observation set. In practice, when the observation space is infinite or too large to handle, we must rely on an empirical estimation of $V$ for decision making, by predicting (sampling) a finite subset of hypotheses for $\mathcal{Z}_{k+1:k+T}$ (for each candidate). 

\begin{figure}[t]
\includegraphics[width=0.9\textwidth]{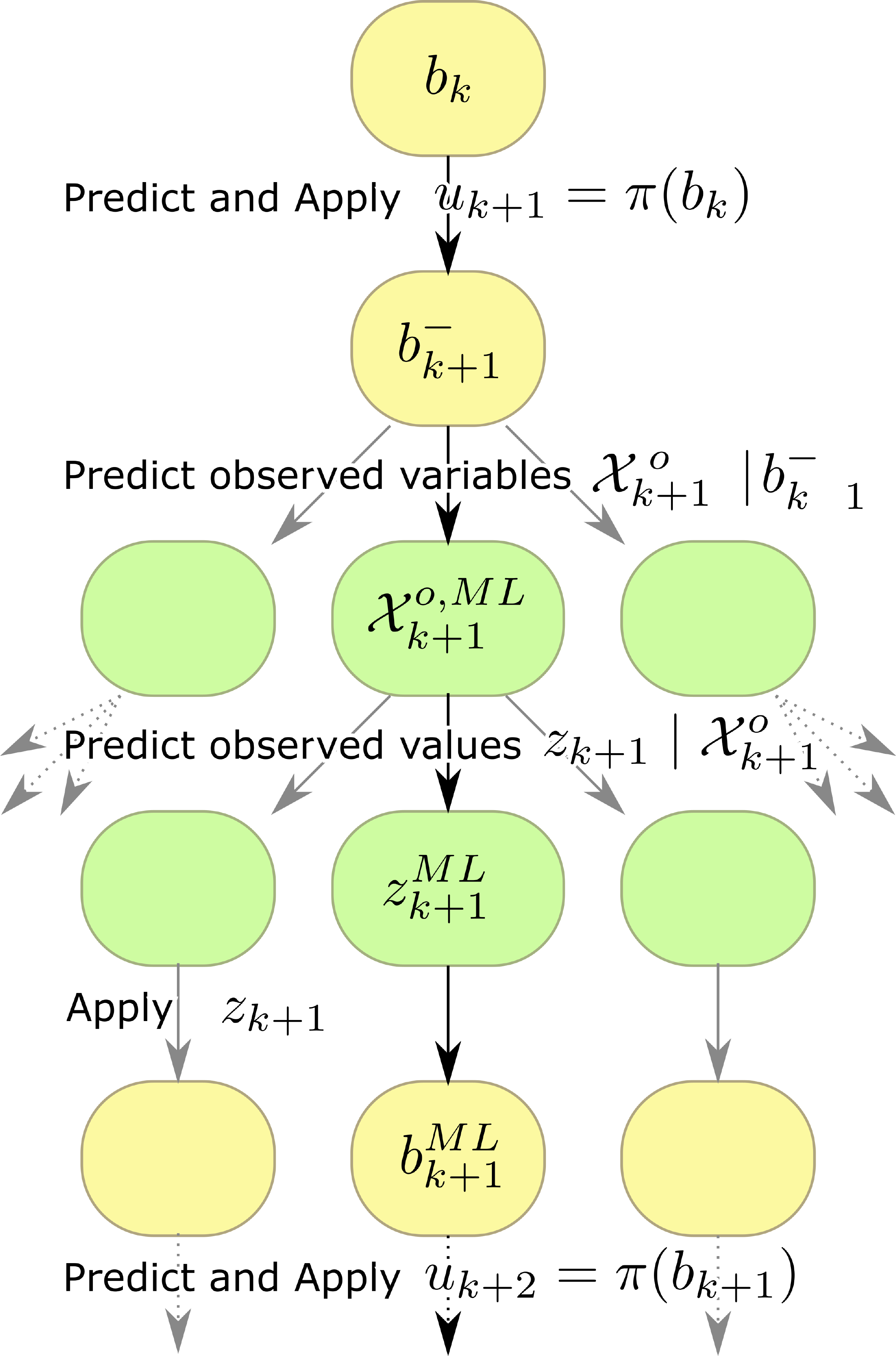}
\caption{The tree of predictable hypotheses of a policy $\pi$, starting from belief $b_k$. Though belief transition is deterministic, uncertainty in the future observations leads to hypothesis branching. The "maximum likelihood" hypothesis (in black) is a specific path in tree, inferable without explicit belief updates.}
\label{fig:hyp-tree}
\end{figure}

To appropriately predict observation realizations according to the generative model (i.e., sample them according to their likelihood), we can examine an "hypothesis tree", for every candidate policy, as shown in Fig.~\ref{fig:hyp-tree}. When following a certain path from the root down to a leaf, the policy is incrementally executed, and control actions and observations are selected (given the latest belief) and incorporated (into that belief) in alternation. Thus, every path in this tree represents a hypothesis -- an appropriate assignment to the Markov chain $H_{k+1:k+T} \doteq [u_{k+1},z_{k+1},\dots,u_{k+T},z_{k+T}]$. After "completing" each step, we can calculate its reward.

Beyond affecting the high cost of belief propagation in high-dimensional states, the "curse of dimensionality" also plays an important role in cost of predicting observations. Generally, if the state is small (e.g., a single pose), or the entire state is observed each time, then we only need to contemplate about the observation value, considering the observation stochastic noise. However, when the observations may involve only a subset of the state variables, as we considered before, we should ask two questions before predicting each observation: "which variables (if any) are observed?", and (only then) "what is the observed value?".

We thus need to "pass" two layers in the hypothesis tree, in order to sample an observation in each time-step. The branching in the first layer represents variation in the topology of the observation, where all observations derived from each of these branches involve the same subset of state variables; the branching in the second layer represents variation in the observed value. This double branching means that the breadth of the hypothesis tree can quickly "explode" beyond our computation ability. This is especially problematic in our case, since: (1)~we want to solve the planning online; (2)~for each hypothesis, we need to perform high-dimensional belief update (and not, e.g., simple particle propagation), which is computationally demanding; (3)~the process of (realistic) observation generation is typically not trivial; e.g., in vision-based navigation, we would have to calculate landmark observability from our current pose (i.e., which landmarks are in our field of view), considering multiple realizations for the state value.

As mentioned, it was shown that using only a single hypothesis per candidate is often enough to achieve sufficient accuracy, by assuming "Maximum Likelihood (ML)" observations (see \cite{Platt10rss}). According to this assumption, we should only consider the "most likely" observation $z_{k+t}$ in each time-step~$k+t$. Such observation relies on the MAP estimate of the belief, which is, in the Gaussian case, its mean $\bm X^*_k$. Given this realization for the state variables, we can determine which variables would be observed (involved) $\mathcal{X}^o_{k+t}$, and infer the ML observation from the observation model. When the model noise is a zero-mean Gaussian, as we considered before, this observation would simply be $z^{ML}_{k+t} = h_{k+t}(\mathcal{X}^o_{k+t})$. So, the "ML observations assumption" translates every candidate policy into a specific "action and observation sequence" (i.e., a single hypothesis $H_{k+1:k+T}$ from the hypothesis tree), according to which the initial belief should be propagated.

Further, it is clear that updating the belief $b^-_{k+t}$ (derived from the belief $b_{k+t-1}$ after performing the action $u_{k+t}$, but before incorporating the observation $z_{k+t}$) using the ML observation $z^{ML}_{k+t}$, would not change its MAP estimate -- since we used this exact value to generate the observation. Thus, assuming the policy function depends only on the belief's MAP estimate, the next action to be taken $u_{k+t+1} = \pi(b_{k+t})$ can be determined without calculating the updated belief $b_{k+t}$, but just using the MAP estimate of $b^-_{k+t}$. Also, updating $b_{k+t}$'s mean, after applying the action $u_{k+t+1}$ (which has a zero-mean noise model), is done by simply calculating the noiseless transition function: $\overline{x_{k+t+1}} = g_k(\overline{x_{k+t}}, u_{k+t+1})$; thus, we can conveniently generate the next ML observation $z^{ML}_{k+t+1}$, without actually calculating the updated belief $b_{k+t+1}$. Overall, this means that the entire "ML hypothesis" (and the posterior MAP estimate) is known \emph{in advance}, before performing any belief updates, and by only iteratively applying the deterministic transition and observation functions.

\subsection{Utilizing PIVOT with General Observations}
As explained, to determine the $\pivot$ order, we need access to the predicted updates, in order to determine the variable involvement levels and classification. We recall that assuming ML observations contributes to planning in two manners: (1) each candidate policy is represented with only a single hypothesis, and can be evaluated with a single update; (2) this hypothesis (along the entire planning horizon) can be inferred ahead of planning, without propagating the belief. Both of these properties are beneficial for $\pivot$, as they allow us to infer the variable involvement levels in the long-horizon future updates, ahead of planning time; in other words, they allow us to "prepare" in advance for all the updates due in the planning process. 

\begin{figure}[b]\center
		\includegraphics[width=0.8\textwidth]{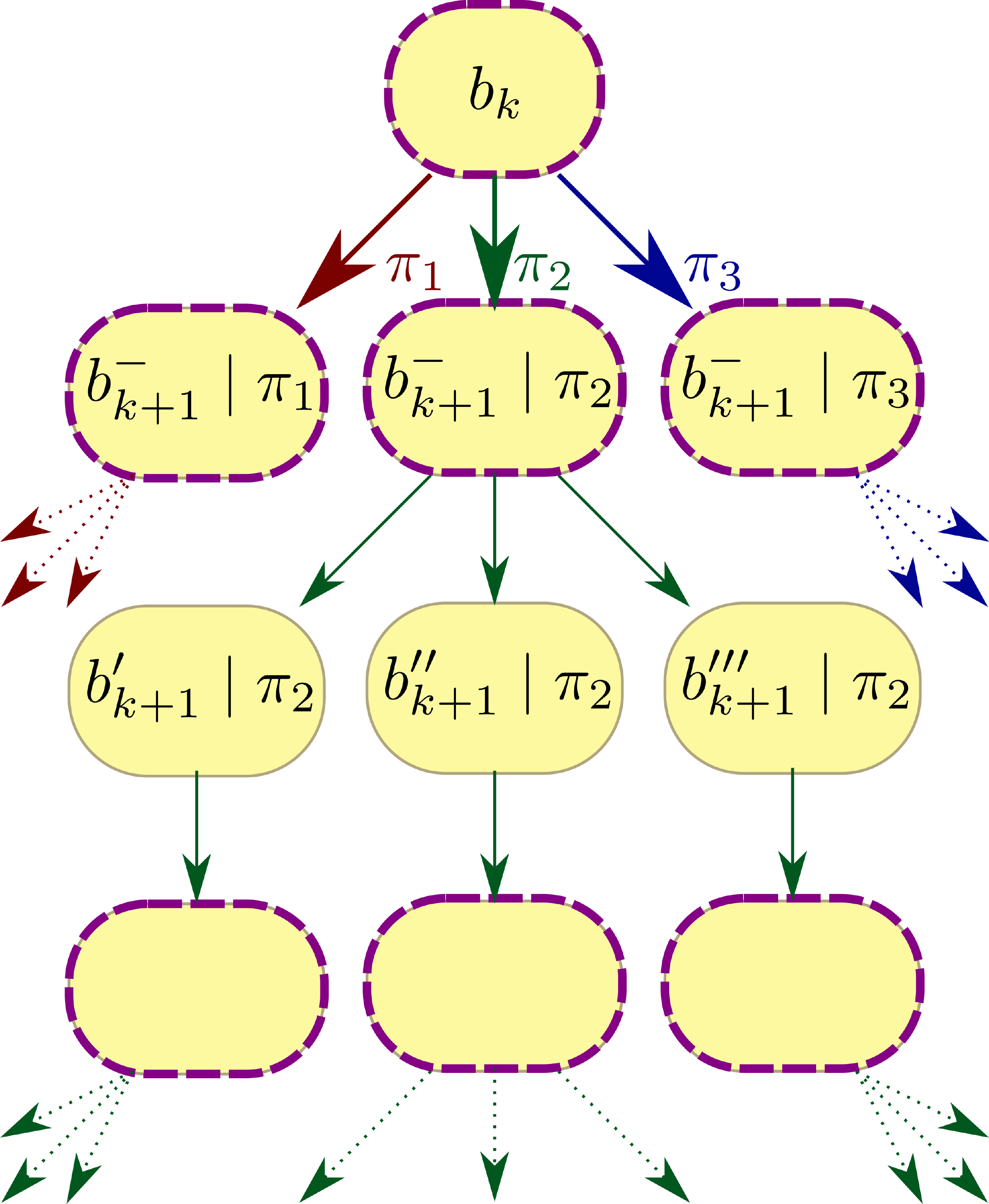}
\caption{Planning trees, built by connecting the full hypothesis trees (as presented in Fig.~\ref{fig:hyp-tree}) of three candidate policies at their root. $\pivot$ is applied on the beliefs at each time-step (in purple), given only the myopic updates. It is also possible to apply $\pivot$ on the initial belief, considering \emph{heuristic} knowledge on the variable involvement levels.}
		\label{fig:planning-tree-general}
\end{figure}

When not relying on ML observations, each candidate is matched with multiple hypotheses. Further, for each candidate, and at each time-step, we have access only to the (multiple) immediately upcoming updates, taking the belief to the next time-step. Meaning, we do not have access to the "complete" hypotheses (over the entire horizon) in advance, but must "build" them gradually, by actually performing the belief updates. Hence, we also cannot infer the involvement levels of the variables in these hypotheses ahead of time, but only infer the involvement levels in the upcoming myopic updates. Thus, in this case, we shall use $\pivot$ myopically and incrementally, by calculating and applying a $\pivot$ order \emph{on the intermediate beliefs} at each time-step. In other words, we shall apply $\pivot$ before each "observation branching" of the hypothesis tree, as demonstrated in Fig.~\ref{fig:planning-tree-general}.


\subsubsection{PIVOT with Heuristic Classification}
Previously, we relied on the "true" variable involvement levels, derive from the (accessible) future updates, to determine $\pivot$. Yet, it is also possible to try and heuristically \emph{estimate} the variable involvement levels, even if we do not have full access to the future updates in advance. For example, we can use the "ML hypotheses" just to estimate the variable involvement, even if we consider planning with multiple hypotheses per candidate. We can also try to perform a precursory visibility or reachability analysis, to identify variables that are (almost) certainly involved or uninvolved in the future updates. In some cases, we may even know in advance that a certain subset of the state variables can never be involved by definition; e.g., past poses are never involved in landmark-based SLAM -- where only landmarks can be re-observed.

In this case, we shall infer and apply a $\pivot$ order on the initial belief, in preparation for the long-horizon future updates, based on this estimated classification. We note that even if this variable classification does not match the "true" classification, inferred from the "real" updates (when they become available), this does cannot lead to any sacrifice in accuracy; at worst, such misclassification can lead to an increase in the computational cost of the updates.

%% file: appendix-models.tex
\section{\changed{Generalization to non-Linear/Gaussian Models\label{sec:appendix-general-models}}}
When considering general models (which are not linear and Gaussian), the problems of state inference and belief space planning can no longer be described using the matrix-based formulation provided in Sections~\ref{sec:prelim-square-root-sam}-\ref{sec:prelim-planning}. Nonetheless, they can still be described using graphical models, as we began to explain in Section~\ref{sec:prelim-infer}; thus, let us now extend that graph-based problem formulation, explain its equivalence to the matrix-based formulation, and show that $\pivot$ is still relevant and applicable in this case.

\subsection{\changedRemoved{Inference with General Beliefs}}
To perform state inference with general models, we can re-factorize the belief $b_k(\mathcal{X}_k)$ (from (\ref{eq:belief-factor-product})) to a product of $\abs{\mathcal{X}_k}$ conditional factors, representing the conditional probabilities of each state variable, given (a subset of) the other variables. This process is known as \emph{variable elimination}. While the order of state variables used in (\ref{eq:belief-factor-product}) (and the factor graph) is arbitrary, the variable order now plays an important role, as it conveys the elimination order chosen for the variables. Let us consider a certain variable order, represented with the state vector $\bm X_k$. Then, in short, this elimination process is done as follows: we choose the first variable for elimination $\bm X_k(1)$, and select all the factors in $\mathcal{F}_k$, in which it is involved. Then, we utilize the chain rule to replace the selected factors with two "new" factors: (a) the conditional probability of $\bm X_k(1)$, given all the other variables involved in the selected factors $\prob(\bm X_k(1) \mid d(\bm X_k(1)))$; and (b) a "marginal factor" $f^\text{marginal}_{k,1} \doteq \prob(d(\bm X_k(i)))$, which connects all these variables (not including $\bm X_k(1)$). We then repeat this process according to the elimination order, examining at each step the remaining factors from $\mathcal{F}_k$, and the accumulated marginal factors. Finally, we shall be left with a product of $n_k$ conditional factors:
\begin{equation}
\label{eq:belief-conditionals-product}
b_k(\bm X_k) \propto \prod_{i=1}^{n_k-1} \prob(\bm X_k(i) \mid d(\bm X_k(i)))\cdot\prob(\bm X_k(n_k)),
\end{equation}
where $d(x)$ denotes the set of variables $x$ depends on (for the chosen variable  order). Surly, a variable may only depend on those which follow it according to the chosen order. Then, from this hierarchical representation, we can derive the MAP estimate (or other marginal distributions), by examining the variables in the reverse order.

\begin{figure}[t]
	\begin{subfigure}[t]{\textwidth}\center
		\includegraphics[width=0.75\columnwidth]{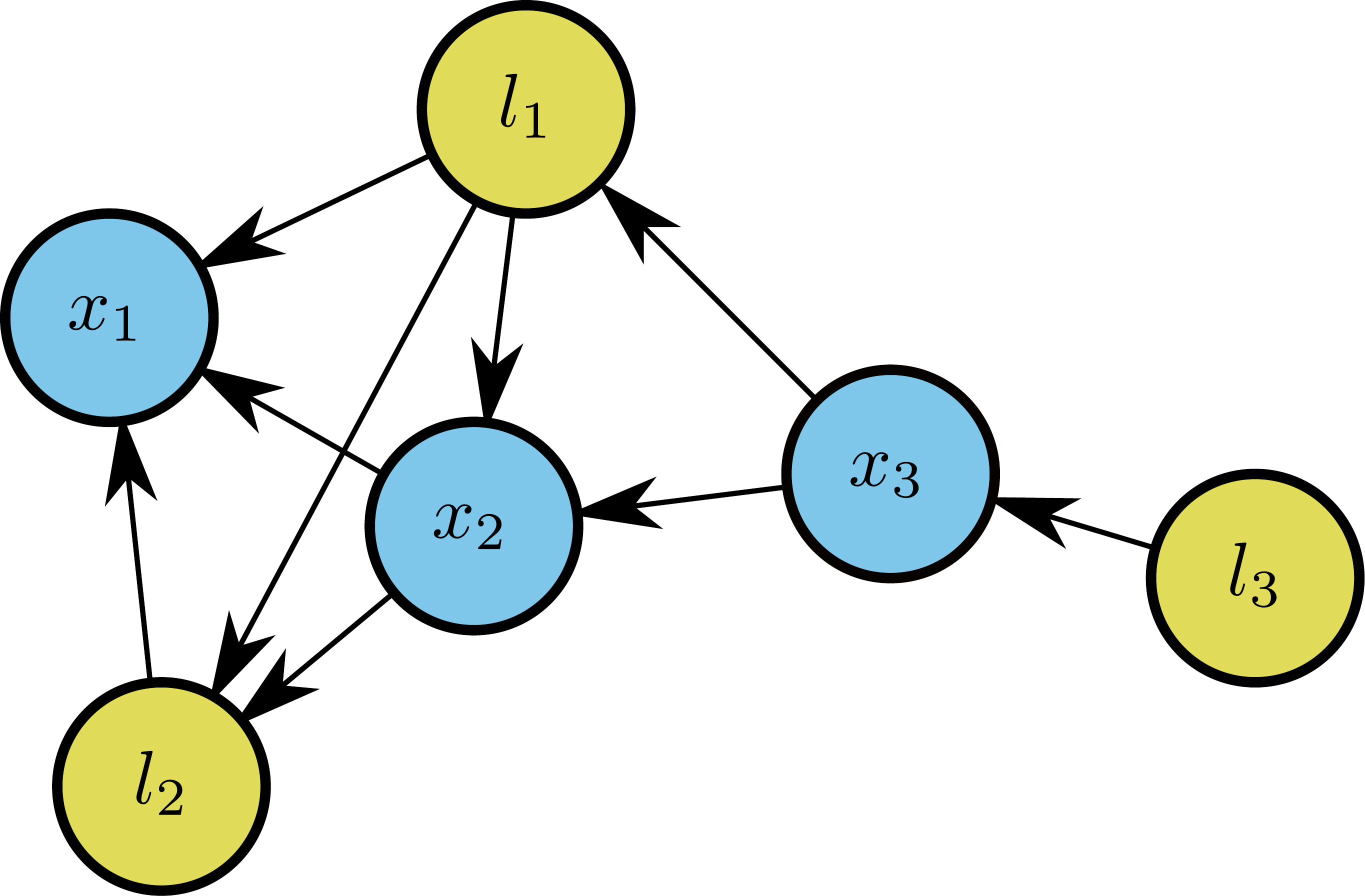}
\caption{The original Bayes net created from elimination of the variables in the factor graph given in Fig.~\ref{fig:example-factor-graph}; this process depends on the elimination order, which here is chosen to be $[x_1,l_2,x_2,l_1,x_3,l_3]$. Arrows between the variable nodes represent conditional dependencies.\\$ $}
		\label{fig:example-bayes-net}
	\end{subfigure}
	\begin{subfigure}[t]{\textwidth}\center
		\includegraphics[width=0.75\columnwidth]{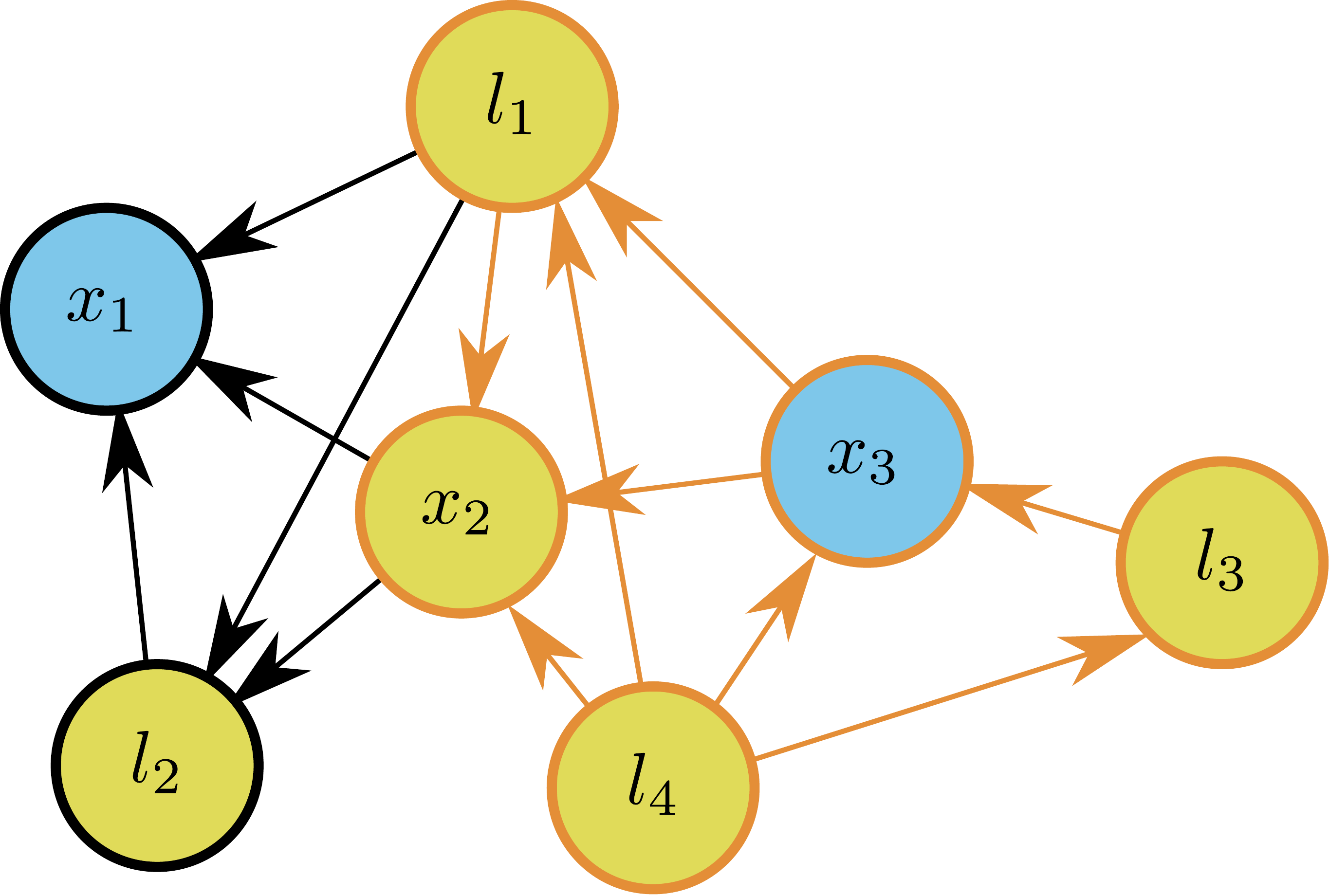}
		\caption{Updated Bayes net, after addition of new factors and/or variables to the factor graph, as shown in Fig.~\ref{fig:inc-update}. The only variables affected by the update (in orange) are the first involved variable, and those which follow it, according to the elimination order.}
		\label{fig:example-bayes-net-update-full}
	\end{subfigure}
\caption{Incrementally updating the Bayes net (i.e., belief factorization), used for inference in non-linear/Gaussian beliefs.}
\label{fig:example-bayes-net-update}
\end{figure}

Graphically, this factorization of the belief to conditional probabilities can be represented with yet another PGM -- a Bayesian network ("Bayes net") $BN_k \doteq (\mathcal{X}_k,\mathcal{D}_k)$. In the Bayes net, the directed edges $\mathcal{D}_k$, represent the conditional dependencies between the connected variables, such that the edge's destination depends on its origin. Elimination can be seen as building the belief's Bayes net from its factor graph. A Bayes net, resulting from eliminating the variables of the factor graph from Fig.~\ref{fig:example-factor-graph}, is provided in Fig.~\ref{fig:example-bayes-net}.

We note that there is a duality between the factorization of $b_k(\bm X_k)$ from a product of $\abs{\mathcal{F}_k}$ factors, to a product of $n_k$ conditional probabilities, and the QR factorization of $\bm J_k$ ($\abs{\mathcal{F}_k}$ rows) to $\bm R_k$ ($n_k$ rows), as described before for linear(ized) Gaussian models. Each conditional probability corresponds (in order) to the respective row of $\bm R_k$, and the sparsity pattern of this matrix matches the structure of the Bayes net $BN_k$: the variable at index~$i$ is dependent on the variable at index~$j$, if and only if ${{\bm R}_k}_{ij}$ is non-zero. Appropriately, edges between variables in the Bayes net, which did not have a connecting factor in the Factor graph, are also known as "fill-in". The amount of fill-in reflects the computational cost of elimination, and depends on the elimination order. For more details, see, e.g., \cite{Elimelech21ral}.

\subsection{\changedRemoved{Incremental Belief Update}}
We recall that we consider a sequential problem, in which the system (i.e., sets of variables and factors) may grow as time progresses. Hence, at the arrival of new factors, we wish to update the belief accordingly, and refine our estimate. 

Formally, we may consider the initial belief $b_k(\mathcal{X}_k)$, represented with $FG_k$ and $BN_k$, and assume that at time-step $k+1$ we wish to update the system with new factors, and possibly new state variables (e.g., transition to a new pose). Such an update can be represented as a factor \mbox{(sub-)graph}, which we wish to merge into the initial graph; i.e., 
$FG^\delta_{k+1} \doteq (\mathcal{X}_k\cup\mathcal{X}^\delta_{k+1}\cup\mathcal{F}^\delta_{k+1},\mathcal{E}^\delta_{k+1})$, such that $ FG_{k+1} \doteq FG_k \cup FG^\delta_{k+1} $, as demonstrated in Fig.~\ref{fig:inc-update}.

Adding new factors to the belief (i.e., updating (\ref{eq:belief-factor-product}) or the matching factor graph) is trivial; yet, inferring the updated factorized belief representation at time $k+1$ (i.e., updating (\ref{eq:belief-conditionals-product}) or the matching Bayes net) is not, and requires re-elimination. Conveniently though, and similarly to the square root matrix, this can be performed incrementally, given access to the elimination process of the belief at the previous time-step $k$. Assume that the first variable in $\mathcal{X}_k$, according to the elimination order $\bm X_k$, that is involved in any of the new factors, is $\bm X_k(j)$. Thus, according to the elimination algorithm explained before, instead of full elimination of $b_{k+1}/FG_{k+1}$, we can reuse the conditionals (and marginal factors) calculated at time-step $k$, and only (re)start the elimination from the $j$-th variable. We refer to the variables which are left for re-elimination, i.e., $\bm X_{k+1}(j)$ to $\bm X_{k+1}(n_{k+1})$, as the "affected variables". Of course, to take advantage of such incremental updates, the elimination order of the first $j-1$ variables in $\bm X_{k+1}$ (the "unaffected variables") must match the one set in $\bm X_{k}$ -- otherwise, they would become affected too. A demonstration of this concept is provided in Fig.~\ref{fig:example-bayes-net-update-full}.

We should also mention that by using more advanced graphical structures, such as the Bayes or Junction tree \citep{Kaess10tr}, it is sometimes possible to avoid re-elimination of some of the variables we defined as "affected" (i.e., all the variables following the first involved). Since this aspect is not essential to the discussion, we choose not to over-complicate this formulation, and assume all affected variables are due to re-elimination.

\subsection{\changed{Applying PIVOT}}
As we may easily conclude from the previous discussion, changing the order  of state variables, as suggested by $\pivot$, is relevant also when planning with non-linear/Gaussian beliefs. In this case, the order of variables corresponds to the elimination order used to create the Bayes net, instead of the order of columns used to create the square root matrix. The influence on the affected variables and fill-in remains similar in both the graphical and matrix representations.